\documentclass[pdflatex,sn-mathphys-ay]{sn-jnl}

\usepackage{graphicx}%
\usepackage{multirow}%
\usepackage{amsmath,amssymb,amsfonts}%
\usepackage{amsthm}%
\usepackage{mathrsfs}%
\usepackage[title]{appendix}%
\usepackage{xcolor}%
\usepackage{textcomp}%
\usepackage{manyfoot}%
\usepackage{booktabs}%
\usepackage{algorithm}%
\usepackage{algorithmicx}%
\usepackage{algpseudocode}%
\usepackage{listings}%
\usepackage{natbib}

\theoremstyle{thmstyleone}%
\newtheorem{theorem}{Theorem}
\newtheorem{proposition}[theorem]{Proposition}%

\theoremstyle{thmstyletwo}%
\newtheorem{remark}{Remark}%

\theoremstyle{thmstylethree}%
\newtheorem{lemma}{Lemma}

\begin{document}

\title[Article Title]{Certified Interpolation Oversampling: Per-Instance Safety Guarantees for Imbalanced Learning}

\author[1]{\fnm{Pankaj} \sur{Yadav}} \email{yadav.58@iitj.ac.in}

\author[2]{\fnm{Vivek} \sur{Vijay}} \email{vivek@iitj.ac.in}

\affil[1,2]{\orgdiv{Department of Mathematics}, \orgname{Indian Institute of Technology Jodhpur}, \orgaddress{\postcode{342030}, \state{Rajasthan}, \country{India}}}








\abstract{

Synthetic minority oversampling is typically designed and evaluated against a predictive objective, generating samples that improve downstream classification. This paper pursues a second objective by generating samples that carry a stated safety property, established for each instance by construction rather than assumed. We introduce Certified Interpolation Safe Oversampling (CISO), a three-phase interpolation framework built for this objective. A safety-guided distribution selects where synthesis occurs, a locality-and clearance-weighted distribution selects with whom each anchor interpolates, and a q-Gaussian placement density determines how far along the resulting segment each instance is placed. The framework provides three guarantees. Each synthetic instance seeded by a sufficiently safe anchor carries a certified distance from the majority class; a signed temperature parameter provably and monotonically shifts synthesis between boundary-seeking and interior-seeking regimes; and selection weights are strictly positive by construction, so no degenerate case arises. Certification is obtained alongside competitive predictive performance rather than in place of it. Under a protocol preregistered before evaluation, across 45 datasets, four classifiers, and eleven competing methods, CISO is statistically equivalent to SMOTE on precision-recall AUC, ranks second of eleven under gradient boosting, and completes every one of 11,460 fold-level evaluations without failure. A parameter sweep further reveals a continuous fidelity-safety trade-off that competing methods occupy only as isolated points.
}

\keywords{Imbalanced learning, Oversampling, Synthetic data, Certified generation, Preregistered evaluation}


\maketitle
\section{Introduction}\label{sec1}

Class imbalance arises whenever the class of interest is severely underrepresented, for example in fraud detection, medical diagnosis, and fault identification. Standard classifiers implicitly assume balanced training data \citep{ci1,ci2}, so this imbalance degrades their performance. Among the remedies, data-level oversampling remains the most widely used~\citep{q11, q17, q6}. It is classifier-agnostic and simple to deploy. The synthetic minority oversampling technique (SMOTE) \citep{ci3} and its interpolation-based descendants dominate current practice~\citep{q8,q13,q23}.

Synthetic minority generation can serve two distinct goals, though the literature has primarily pursued only one of them. The first goal is predictive, which involves generating samples that help a downstream classifier separate the classes. This is the goal most of the interpolation literature has been designed and evaluated against \citep{q4,q16}. The second goal is to generate samples with a stated, verifiable safety property, ensuring a synthetic minority instance does not land inside the majority region. Critically, this property should be certified for each instance rather than assumed, while remaining competitive on the predictive goal rather than trading it away. Per-instance evaluation of generated data is itself a recognized pursuit in the generative modeling literature \citep{ci4}, though existing approaches audit samples after generation rather than building the guarantee directly into the generator. This certification objective becomes critical whenever synthetic data is deployed beyond a single tuned classifier. Resampling choices are known to interact with the downstream model rather than transfer uniformly across it \citep{ci5}. Furthermore, in settings where synthetic instances feed a human-reviewed decision or a post-hoc explanation, resampling can distort the faithfulness of what is presented to the analyst \citep{ci6}. In regulated domains such as healthcare and finance, practitioners and regulatory bodies have called explicitly for auditing and certification mechanisms to verify that synthetic data meets stated ethical and legal standards \citep{ci7}, a requirement a sampler cannot satisfy if its safety properties are assumed rather than computed. No existing interpolation oversampler was built to provide a per-instance certified property of this kind, and none reports one. This paper directly addresses this gap.

Building a sampler for the certification goal requires deciding where its safety information acts. A substantial branch of the predictive-goal literature places such information in the selection weights, injecting local density or distance to steer generation away from majority-dominated regions, a pattern spanning foundational methods and persisting in recent proposals \citep{ci8,ci12,ci13}. That placement carries a constraint. Any design that folds anchor-level information into a candidate weight and subsequently normalizes that weight within a single anchor's neighborhood loses the information it intended to inject, since any component constant across the neighborhood cancels identically from the resulting selection probabilities. Anchor-level safety information must therefore act where it varies, in the choice of which anchor seeds synthesis.

We build the framework accordingly. Synthesis proceeds through three distinct phases, each governing a single decision and incorporating a precise piece of geometric information. First, a safety-guided anchor-selection distribution determines \emph{where} synthesis occurs. Next, a locality- and clearance-weighted neighbor distribution decides \emph{with whom} each anchor interpolates. Finally, a stochastic interpolation step, driven by a q-Gaussian placement density, determines \emph{how far} along the resulting segment each synthetic instance is placed.

The framework carries three guarantees that, to our knowledge, no existing interpolation oversampler provides. Every synthetic instance seeded by an anchor with positive certified clearance provably lies at least a computable distance from every majority instance. A single signed temperature provably and monotonically shifts synthesis between borderline-seeking and interior-seeking regimes, recovering the two opposing philosophies of the literature within one formula. Furthermore, because selection weights are strictly positive by construction, no degenerate case exists.

We test these guarantees directly, under a preregistered protocol across 37 KEEL benchmark datasets, 8 larger UCI datasets, and four classifiers. The framework achieves its guarantees while remaining competitive on predictive performance. Specifically, it is statistically equivalent to SMOTE on the primary threshold-free metric, ranks second of eleven methods under gradient boosting, and significantly outperforms two established density-aware baselines. Separately, no sophisticated oversampler in the comparison, including our own, consistently outperforms class
weighting or random oversampling. This finding is not specific to our method as it extends, under preregistration, a conclusion the recent literature has been converging toward \citep{ci14,ci15}. We report every preregistered outcome in full, including those that did not favor our initial hypotheses.

The reliability and monotonicity guarantees are also confirmed empirically. Across approximately fifty thousand evaluations, at dimensionalities up to 166 and sample sizes up to 20{,}000, the framework completed every run without a single sampler failure, whereas some established density-aware baselines failed on most or all datasets under the same protocol. An exploratory sweep of the temperature parameter confirms the monotonicity guarantee directly. Measured clearance rises monotonically exactly as predicted, while held-out distributional fidelity reaches its maximum at the neutral setting and
decreases as synthesis is reweighted toward safety. This reveals that the fidelity-safety trade-off is a continuous choice rather than a static algorithm design feature, whereas existing methods occupy only a single fixed point along that continuum.

The contributions of this paper are as follows:
\begin{itemize}
\item We construct a three-phase interpolation oversampling framework that provides, to our knowledge, the first per-instance certified clearance guarantee in this family, together with a provably monotone control parameter and well-posedness across all edge cases. 
\item We test whether these guarantees hold empirically in accordance with our theoretical predictions, under a fully preregistered evaluation across 45 datasets, four classifiers, and eleven methods.
\item Our results locate existing interpolation oversamplers on the fidelity-safety trade-off this reveals, showing that our control parameter traverses this spectrum continuously where competing methods occupy only fixed points.
\end{itemize}

The remainder of the paper is organized as follows. Section~\ref{sec:related} reviews related work. Section~\ref{sec:prelim} fixes notation and preliminaries. Section~\ref{sec:method} develops the CISO framework and its guarantees. Section~\ref{sec:protocol} specifies the experimental protocol, Sect.~\ref{sec:results} reports the results, and Sect.~\ref{sec:discussion} discusses implications and concluding remarks.


\section{Related Work}\label{sec:related}

Handling imbalanced data has been studied extensively, and data-level resampling remains the most widely adopted family of remedies~\citep{q5,q15}. SMOTE~\citep{kan42} generates synthetic minority instances by linear interpolation between a minority instance and a randomly chosen minority neighbour. Its simplicity and classifier-agnosticism design have made it the foundation of a broad family of extensions. A recurring theme in that family is the replacement of SMOTE's uniform choices with geometry-aware mechanisms. However, the specific point in the generation pipeline at which this geometry is applied varies across methods. ADASYN~\citep{EC5} allocates more synthetic instances to minority instances surrounded by majority neighbours, normalizing a difficulty ratio across the minority class to decide how many samples each instance seeds. Borderline-SMOTE~\citep{ci9} applies a
hard filter, restricting synthesis to instances near the decision boundary and interpolating uniformly thereafter. Safe-Level-SMOTE~\citep{ci10} leaves anchor selection unchanged but biases placement along the interpolation segment according to a ratio of minority neighbour counts. MWMOTE~\citep{ci11} weights hard-to-learn instances by their distance to the majority class before resampling them through a clustering step. Gaussian Distribution based
Oversampling \citep{ci16} selects anchors by roulette-wheel draw over a combined density and distance weight. It then displaces each anchor along a uniformly random direction by a Gaussian-distributed radius rather than interpolating between existing instances. Recent proposals continue in this direction, injecting local density or
adaptive weighting into the selection step~\citep{q20,q21A}. A separate line removes the
neighbourhood size parameter entirely, defining neighbourhoods through mutual nearest-neighbour relations that stabilize adaptively~\citep{ci17, ci18}.

Alongside these constructive proposals, a smaller body of work analyzes oversampling rather than extending it. Elreedy and Atiya~\citep{ci19} characterize the expectation and covariance of SMOTE-generated instances, and a subsequent analysis~\citep{ci20} derives the full probability density of SMOTE patterns, showing that synthetic instances do not in general follow the minority class distribution. Moniz and Monteiro~\citep{ci15} apply no-free-lunch reasoning to imbalanced learning. Furthermore, a large systematic study \citep{ci14} reports that with well-tuned strong classifiers, SMOTE and its variants rarely improve over class weighting or trivial resampling on threshold-free metrics. These analytical insights motivate both the evaluation protocol adopted here and the choice of PR-AUC as the primary metric.

A different concern arises when synthetic data is evaluated at the level of the individual instance rather than the population. Sample-level metrics have been developed to audit the fidelity and authenticity of samples a generative model has already produced, and to discard low-quality ones after the fact \citep{ci4}. However, such auditing is post-hoc, applied to a black-box generator's output rather than built directly into the generation process. Approaches that estimate the minority density directly and sample from it~\citep{ci21,ci22}, or that employ deep generative models for high-dimensional data~\citep{ci23}, target a fundamental resource regime. Density estimation is inherently unreliable when minority samples are scarce and generative models require training budgets that far exceeds those of the standard tabular benchmarks considered here.

Two gaps follow from this survey. First, the geometry-aware methods above apply their structural information at different pipeline stages, and the exact stage at which a quantity is applied determines whether it can influence the decision it is meant to inform (Sect.~~\ref{sec:method}). Second, none of these methods states a per-instance safety property that holds by construction at generation time. For instance, GDO can place synthetic instances inside the majority region when assigned an overly wide generation radius~\citep{ci16}, while sample-level auditing addresses this concern only after generation has occurred. The framework developed here computes the property from the training geometry before generation, ensuring that the guarantee holds for every instance produced.

\section{Preliminaries}\label{sec:prelim}

This section fixes notation and introduces the mathematical background the framework draws on, the $k$-nearest-neighbor density estimator that grounds its density-based scores, and the q-Gaussian kernel it uses directly. Constructions specific to the framework are deferred to Sect.~\ref{sec:method}.

\subsection{Problem setup and notation}\label{subsec:setup}
Let $D=\{(x_i, y_i)\}_{i=1}^N$ be a binary-labeled dataset with feature vectors $x_i \in \mathbb{R}^d$ and labels $y_i \in \{0,1\}$, where $y_i=1$ denotes the minority class. Features are standardized to zero mean and unit variance, so that all distance metrics share a common scale. Define $D_{\text{maj}} = \{x_u^{\text{maj}}\}_{u=1}^{N_0}$ and $D_{\text{min}} = \{x_i^{\text{min}}\}_{i=1}^{N_1}$ for the majority and minority subsets, with $N_0 > N_1 \ge 2$. The objective is to generate $\Delta N = N_0 - N_1 > 0$ synthetic minority instances so that the augmented dataset is class-balanced.

All geometric quantities are defined with respect to a metric $\ell(\cdot,\cdot)$ on $\mathbb{R}^d$; throughout this work, we adopt the standard Euclidean metric $\ell(x,x') = \lVert x - x'\rVert_2$. Notably, our theoretical results rely solely on the metric properties of $\ell$ (in particular, the triangle inequality); thus alternative metrics may be substituted without altering the underlying theory.

\subsection{Nonparametric density estimation}\label{subsec:foundation}
The framework's density-based quantities are monotone surrogates for a nonparametric density estimate rather than calibrated probabilities. For a sample of size $n$, the $k$-nearest-neighbor ($k$-NN) density estimator~\citep{ci24} is
\begin{equation}
\hat{p}_k(x) = \frac{k}{n\,V_d\,R_k(x)^d},
\label{eq:knn_estimator}
\end{equation}

where $R_k(x)$ is the distance from $x$ to its $k$-th nearest sample point and $V_d$ is the volume of the unit ball in $\mathbb{R}^d$. Unlike a fixed-bandwidth (Parzen) estimator, the $k$-NN estimator adapts its effective bandwidth $R_k(x)$ to the local sample density, a property that matters under severe class imbalance, where the majority density varies by orders of magnitude between dense cores and the sparse regions adjoining the minority manifold. Because the framework uses density only through bounded, order-preserving scores, any
strictly monotone transformation of Eq.~\eqref{eq:knn_estimator} is equivalent for our purposes; we use inverse mean neighbor distance, which preserves the ordering while avoiding the numerical degeneracy of $R_k(x)^d$ in moderate dimension.

\subsection{The q-Gaussian kernel}\label{subsec:qgaussian}
The framework uses a single kernel in two roles, applying it first to neighbor weighting and then to interpolation placement. For $q > 1$, the q-Gaussian kernel~\citep{ci26} is
\begin{equation}
G_q(t) = \big[\,1 + (q-1)\,t^2\,\big]^{\frac{1}{1-q}}, \qquad t \ge 0.
\label{eq:qkernel}
\end{equation}
It is strictly positive and strictly decreasing in $|t|$, recovers the
Gaussian $e^{-t^2}$ as $q \to 1^{+}$, and exhibits polynomial (heavy) tails
$G_q(t) \sim \big((q-1)t^2\big)^{1/(1-q)}$ as $t \to \infty$. The single
parameter $q$ thus interpolates between light-tailed, strongly local
behavior and heavy-tailed, exploratory behavior; both roles in
Sect.~\ref{sec:method} exploit this control.

\subsection{Neighborhood construction and extremal statistics}\label{subsec:neighborhoods}
For each minority anchor $x_i^{\text{min}}$, define the ordered
neighborhood tuples
\begin{align}
\mathcal{M}_k(i) &= \big(m_{i1}^{\text{maj}}, \dots, m_{ik}^{\text{maj}}\big),
\label{eq:maj_neighborhood}\\
\mathcal{N}_k(i) &= \big(n_{i1}^{\text{min}}, \dots, n_{ik}^{\text{min}}\big),
\label{eq:min_neighborhood}
\end{align}
where $\mathcal{M}_k(i)$ contains the $k$ nearest~\citep{q36} majority instances to
$x_i^{\text{min}}$ and $\mathcal{N}_k(i)$ its $k$ nearest minority
instances \emph{excluding the anchor itself}, each ordered by nondecreasing
distance under $\ell$ with ties broken by dataset index. We require
$1 \le k \le \min(N_1-1,\,N_0)$, truncating $k$ to this range when
necessary. The ordering serves only to define two extremal statistics,
\begin{equation}
c_i = \ell\big(x_i^{\text{min}}, m_{i1}^{\text{maj}}\big), \qquad
r_i = \ell\big(x_i^{\text{min}}, n_{ik}^{\text{min}}\big),
\label{eq:extremal}
\end{equation} 
the \emph{anchor clearance} $c_i$ (distance to the nearest majority instance) and the \emph{neighborhood radius} $r_i$ (distance to the farthest of the $k$ minority neighbors). No correspondence between equally-indexed elements of $\mathcal{M}_k(i)$ and $\mathcal{N}_k(i)$ is implied or used; this point becomes essential in Sect.~\ref{sec:method}.

\section{The CISO Framework}\label{sec:method}

We present Certified Interpolation Safe Oversampling (CISO). Synthesis proceeds in three phases, and the section is organized so that each phase is preceded by the mathematical objects it requires and followed by the formal properties it guarantees. Before presenting the construction, we state the informal guarantee that motivates it. \emph{Every synthetic instance CISO generates from a sufficiently safe anchor provably lies at least a computable distance from every majority instance, and a single parameter shifts synthesis between boundary-seeking and interior-seeking regimes in a provably monotone way.} The certified-clearance guarantee is made precise in Lemma~\ref{lem:margin} and the monotonicity in Proposition~\ref{prop:monotone}, once the formulations they refer to has been defined.

\subsection{Density field and safety score}\label{subsec:field}

CISO concentrates all majority-density and clearance information in
\emph{anchor-level} quantities. This choice is not incidental; selection
weights in interpolation oversampling are normalized within each anchor's
neighborhood, and any factor constant across that neighborhood cancels from
the resulting probabilities. Anchor-level information therefore cannot act
through neighbor weights and must instead govern anchor selection. We prove
this formally in Sect.~\ref{subsec:anchor} (Lemma~\ref{lem:cancellation}) and
here construct the anchor-level quantities it dictates.

\paragraph{Majority density field.}
All density information is computed once, as a field over the majority class,
and read off by each anchor. For each majority instance $x_u^{\text{maj}}$,
let $\bar{d}_u$ be the mean distance to its $k$ nearest majority neighbors
(excluding itself); by Eq.~\eqref{eq:knn_estimator} this is a monotone
surrogate for the local majority density at $x_u^{\text{maj}}$. Define
\begin{equation}
\rho_u^{\text{raw}} = \frac{1}{\bar{d}_u + \varepsilon}, \qquad
\hat{\rho}_u =
\frac{\rho_u^{\text{raw}} - \min_r \rho_r^{\text{raw}}}
     {\max_r \rho_r^{\text{raw}} - \min_r \rho_r^{\text{raw}}}
\;\in\; [0,1],
\label{eq:density_field}
\end{equation}

with $\varepsilon > 0$ guarding against coincident points, $r$ ranging over
all majority instances, and the convention $\hat{\rho}_u \equiv \tfrac{1}{2}$
when the raw densities are constant. The vector $\hat{\rho}\in[0,1]^{N_0}$ is
the \emph{majority density field}; larger $\hat{\rho}_u$ marks locally denser,
more overlap-prone regions. We use a local $k$-NN mean rather than a global
mean intra-class distance, since the latter conflates local density with
centrality. Each anchor reads the field over its majority neighborhood,
giving the \emph{regional majority density}
\begin{equation}
\rho_{\text{Reg}}(i) = \frac{1}{k}\sum_{p=1}^{k}
\hat{\rho}_{\,m_{ip}^{\text{maj}}} \;\in\; [0,1].
\label{eq:rhoreg}
\end{equation}

\begin{figure}
    \centering
    \includegraphics[width=\textwidth]{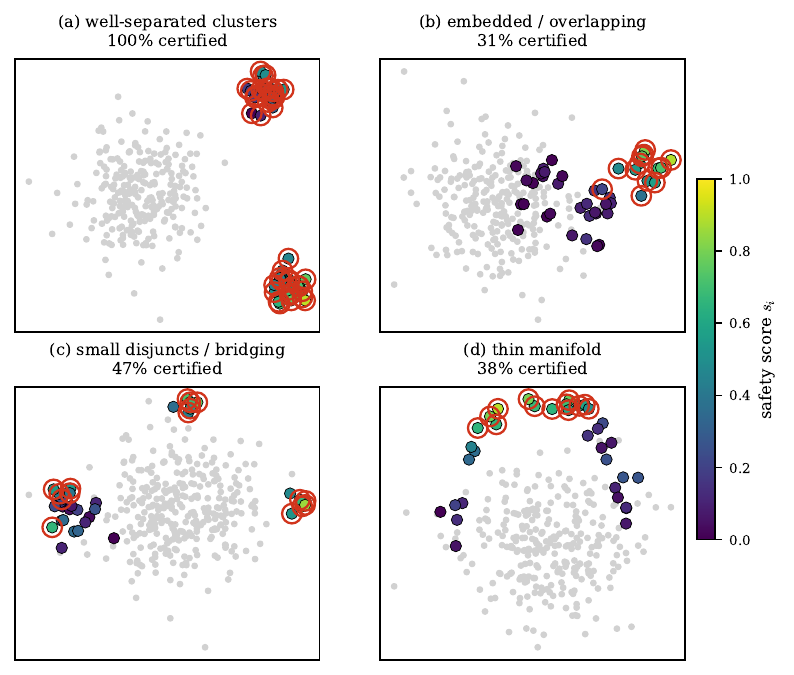}
    \caption{The clearance certificate across four controlled minority geometries. Majority instances are grey; minority instances are coloured by safety score $s_i$; anchors with a positive certificate $\gamma_i = c_i - r_i > 0$ (Sect.~\ref{subsec:guarantees}) are ringed in red. The certified fraction (panel titles) responds to structure: complete for well-separated clusters (a), collapsing under overlap (b), distinguishing a safe cluster from bridging small disjuncts within one dataset (c), and tracking a curved manifold (d). Axes are the standardized features.}
    \label{fig:certificate}
\end{figure}

\paragraph{Anchor safety score.}
Let $\hat{c}_i \in [0,1]$ be the min--max normalization of the clearance
$c_i$ (Eq.~\eqref{eq:extremal}) over the minority set, under the same
constant-case convention. The \emph{safety score} of anchor
$x_i^{\text{min}}$ is
\begin{equation}
s_i = \hat{c}_i\,\big(1 - \rho_{\text{Reg}}(i)\big) \;\in\; [0,1].
\label{eq:safety}
\end{equation}
The product encodes a conjunction; $s_i$ is large only when the anchor is far
from the nearest majority instance \emph{and} faces a sparse majority region.
In the terms of Sect.~\ref{subsec:foundation}, $s_i$ is a two-scale monotone
summary of how weakly the majority class occupies the region around the
anchor. The bounded multiplicative form ensures neither factor dominates and
no reciprocal blow-up occurs; $s_i$ and its constituents lie in $[0,1]$. The
safety score is the sole input to anchor selection. Figure~\ref{fig:certificate} illustrates the safety score and the resulting certificates on four controlled minority geometries.

\subsection{Anchor selection}\label{subsec:anchor}

\paragraph{Where anchor-level information can act.}
We first make precise the principle invoked above.
\begin{lemma}[Within-neighborhood invariance]\label{lem:cancellation}
Fix an anchor $x_i^{\text{min}}$ and suppose the unnormalized selection weight
of its $j$-th neighbor factors as $w_{ij} = a_i\,b_{ij}$, where $a_i>0$
depends only on $i$ and $b_{ij}>0$. Then the normalized probabilities
$\tilde{w}_{ij} = w_{ij}/\sum_{j'=1}^{k} w_{ij'} = b_{ij}/\sum_{j'=1}^{k}b_{ij'}$
are independent of $a_i$.
\end{lemma}
\begin{proof}
$a_i$ is independent of $j$, hence a common factor of numerator and every
denominator term, and cancels.
\end{proof}

Any anchor-level quantity (clearance, regional density, or the safety score
$s_i$ itself) enters a neighbor weight as a factor $a_i$ common to all $k$
candidates and thus cancels; it cannot influence neighbor selection. This is
the sense in which anchor-level density-awareness applied through neighbor
weights is \emph{void}. Consequently CISO uses $s_i$ where it does vary across
the relevant choice; in selecting \emph{which} anchor seeds synthesis. Note that $s_i$ does not cancel in Eq.~\eqref{eq:anchor_dist}, because that normalization runs over anchors rather than within a single anchor's neighborhood: $s_i$ varies across the index being normalized, which is precisely the condition Lemma~\ref{lem:cancellation} requires for a quantity to survive.

\paragraph{The anchor-selection distribution.}
Each synthesis round draws an anchor $I$ from
\begin{equation}
P_\alpha(i) =
\frac{\big(\varepsilon_0 + s_i\big)^{\alpha}}
     {\sum_{r=1}^{N_1}\big(\varepsilon_0 + s_r\big)^{\alpha}},
\qquad \varepsilon_0 > 0,\;\; \alpha \in \mathbb{R},
\label{eq:anchor_dist}
\end{equation}
so the expected number of synthetic instances seeded by anchor $i$ is
$\Delta N \cdot P_\alpha(i)$. The parameter $\alpha$ is a signed temperature
and $\varepsilon_0$ a floor whose roles the next result makes precise.

\begin{lemma}[Properties of $P_\alpha$]\label{lem:palpha}
For all $\varepsilon_0>0$:
\emph{(i)} $P_0$ is uniform on $D_{\text{min}}$;
\emph{(ii)} for $\alpha'>\alpha$, the ratio $P_{\alpha'}(i)/P_{\alpha}(i)$ is
nondecreasing in $s_i$, so $\{P_\alpha\}$ is ordered by likelihood ratio in
$s$, and increasing $\alpha$ shifts anchor mass toward higher safety scores in
the sense of first-order stochastic dominance of $s_I$;
\emph{(iii)} for $\alpha\ge 0$, every anchor satisfies
$P_\alpha(i) \ge \varepsilon_0^{\alpha}\big/\!\big(N_1(\varepsilon_0+1)^{\alpha}\big) > 0$.
\end{lemma}
\begin{proof}
\emph{(i)} At $\alpha=0$ every numerator equals $1$.
\emph{(ii)} $P_{\alpha'}(i)/P_{\alpha}(i)\propto(\varepsilon_0+s_i)^{\alpha'-\alpha}$
is nondecreasing in $s_i$; likelihood-ratio ordering implies first-order
stochastic dominance.
\emph{(iii)} Bound the numerator below by $\varepsilon_0^{\alpha}$ and each of
the $N_1$ denominator terms above by $(\varepsilon_0+1)^{\alpha}$.
\end{proof}

Property~(i) nests uniform (density-agnostic) seeding at $\alpha=0$, so the
contribution of anchor selection is measurable by sweeping $\alpha$ through
zero. Property~(ii) makes $\alpha$ a provably monotone control: $\alpha>0$
concentrates synthesis on safe, interior anchors, while $\alpha<0$ reverses
the ordering and concentrates it near the boundary, recovering
borderline-oriented behavior within the same one-parameter family.
Property~(iii) guarantees, for $\alpha\ge 0$, that no minority instance is
ever excluded from synthesis, with floor controlled by $\varepsilon_0$.

\subsection{Neighbor weighting}\label{subsec:neighbor}

Given an anchor $x_i^{\text{min}}$ drawn in Phase~I, Phase~II selects two of its minority neighbors as the endpoints of interpolation. By Lemma~\ref{lem:cancellation}, the selection weights may depend only on quantities that vary across candidates. Two such quantities suffice, and each is a property of the candidate alone.

\paragraph{Neighbor safety.}
The first is the candidate's own clearance. For neighbor
$n_{ij}^{\text{min}}\in\mathcal{N}_k(i)$, let $\hat{c}(n_{ij}^{\text{min}})$
denote its entry in the minority clearance field. This value is the min--max normalized
distance from $n_{ij}^{\text{min}}$ to its nearest majority instance,
precomputed exactly as $\hat{c}_i$ in Sect.~\ref{subsec:field}. A candidate
close to the majority class is a poor interpolation endpoint, and this term
downweights it.

\begin{remark}[Own clearance versus shared context]\label{rem:confound}
A seemingly natural alternative scores a candidate by its mean distance to the
\emph{anchor's} majority neighborhood $\mathcal{M}_k(i)$. Writing
$\delta_{ij}=\ell(x_i^{\text{min}},n_{ij}^{\text{min}})$ and letting
$\bar{a}_i$ be the anchor's mean distance to $\mathcal{M}_k(i)$, the reverse triangle inequality yields a lower bound for that alternative score that grows at least as $\delta_{ij}-\bar{a}_i$. Consequently, under a safety-seeking sign, this alternative systematically favors remote, potentially cross-cluster candidates, thereby confounding safety with anti-locality. The candidate's own clearance, being a property of the candidate alone, carries no such confound.
\end{remark}

\paragraph{Locality.}
The second quantity is locality, measured on a per-anchor scale and passed
through the q-Gaussian kernel of Eq.~\eqref{eq:qkernel}. With
$\delta_{ij}=\ell(x_i^{\text{min}},n_{ij}^{\text{min}})$, define the
median-normalized distance and the locality weight
\begin{equation}
\hat{d}_{ij} = \frac{\delta_{ij}}{h_i}, \qquad
h_i = \operatorname*{median}_{p=1,\dots,k}\delta_{ip} + \varepsilon,
\label{eq:medscale}
\end{equation}
so that $G_q(\hat{d}_{ij})$ is near-constant when all neighbors lie at
comparable distances (a homogeneous, single-cluster neighborhood). Conversely, the kernel decays sharply for candidates at large multiples of the median distance, which is the signature of an anchor in a small cluster whose nominal neighbors reach across a class boundary. Median normalization makes the kernel self-calibrating, allowing it to adapt to each anchor's own scale without an additional parameter.

\paragraph{Selection weights.}
Combining the two candidate-specific quantities multiplicatively,
\begin{equation}
w_{ij} = \big(\varepsilon_0 + \hat{c}(n_{ij}^{\text{min}})\big)\,
         G_q\!\big(\hat{d}_{ij}\big), \qquad
\tilde{w}_{ij} = \frac{w_{ij}}{\sum_{j'=1}^{k} w_{ij'}} .
\label{eq:phase2_weight}
\end{equation}
A good endpoint is safe \emph{and} local, in the same conjunctive spirit as
the anchor score $s_i$. The following establishes that
Eq.~\eqref{eq:phase2_weight} is always a valid distribution.

\begin{lemma}[Well-posedness and locality control]\label{lem:phase2}
For $q>1$ and $\varepsilon_0,\varepsilon>0$:
\emph{(i)} $w_{ij}>0$ for all $i,j$, so $\tilde{w}_{i\cdot}$ is a well-defined
probability vector for every anchor, with no degenerate case requiring a
fallback;
\emph{(ii)} for candidates of equal clearance, the selection odds reduce to the kernel ratio $\tilde{w}_{ij}/\tilde{w}_{ij'} = G_q(\hat{d}_{ij})/G_q(\hat{d}_{ij'})$. For $\hat{d}_{ij}=\lambda\hat{d}_{ij'}$ with $\lambda>1$, this ratio tends to $\lambda^{-2/(q-1)}$ as $\hat{d}_{ij'}\to\infty$, where the exponent decreases in
q so that smaller values of q enforce stricter locality, whereas larger values permit broader exploration.


\end{lemma}
\begin{proof}
\emph{(i)} Both factors of $w_{ij}$ are strictly positive, the clearance
factor is at least $\varepsilon_0>0$, and $G_q(\hat{d}_{ij})>0$ for $q>1$ and
finite argument. Hence the denominator of $\tilde{w}_{ij}$ is positive.
\emph{(ii)} The common clearance factor cancels in the ratio; the limit
follows from $G_q(t)\sim\big((q-1)t^2\big)^{1/(1-q)}$ as $t\to\infty$
(Eq.~\eqref{eq:qkernel}).
\end{proof}

\paragraph{Sampling the endpoints.}
The two endpoints, denoted $n_{i,a}^{\text{min}}$ and $n_{i,b}^{\text{min}}$,
are drawn from $\tilde{w}_{i\cdot}$ \emph{without replacement}. The first index
$a\sim\operatorname{Cat}(\tilde{w}_{i\cdot})$, and the second
$b\sim\operatorname{Cat}(\tilde{w}_{i\cdot}\mid j\neq a)$ from the renormalized
remainder. Two endpoints, rather than one neighbor as in SMOTE, place the
synthetic instance in the interior of the minority manifold rather than on a
spoke radiating from the anchor's chosen neighbor. Sampling without
replacement guarantees $a\neq b$ whenever $k\ge 2$, and hence a nondegenerate
interpolation segment. It also removes the duplication channel by which weight
concentration would otherwise collapse synthesis onto existing instances,
since under independent draws the coincidence probability
$\sum_j \tilde{w}_{ij}^2$ is largest exactly when the weights are most
concentrated. For the boundary case $k=1$ the single neighbor serves as both
endpoints and Phase~III degenerates to that point.

\subsection{Interpolation}\label{subsec:interp}

Given the anchor $x_i^{\text{min}}$ and the distinct pair
$(n_{i,a}^{\text{min}},n_{i,b}^{\text{min}})$ from Phase~II. Phase~III places
one synthetic instance on the segment joining them. The placement is
\emph{stochastic}, the interpolation coefficient is drawn from a truncated
q-Gaussian density centered where the anchor indicates. Consequently the same, kernel
governing \emph{which} neighbors are eligible in Phase~II also governs
\emph{where along the segment} mass is placed in Phase~III.

\paragraph{Anchor-guided center.}
The center of placement is the metric-only ratio
\begin{equation}
\mu_i = \frac{\ell\big(x_i^{\text{min}}, n_{i,a}^{\text{min}}\big)}
             {\ell\big(x_i^{\text{min}}, n_{i,a}^{\text{min}}\big)
            + \ell\big(x_i^{\text{min}}, n_{i,b}^{\text{min}}\big)}
\;\in\;[0,1],
\label{eq:center}
\end{equation}
so that placement leans toward whichever endpoint lies nearer the anchor whose
safety Phase~I vetted. The anchor thereby participates geometrically, not
merely through selection.

\paragraph{Coefficient distribution and synthesis.}
The interpolation coefficient $\lambda\in[0,1]$ is drawn from the truncated
q-Gaussian density
\begin{equation}
f_i(\lambda) =
\frac{G_q\!\big((\lambda-\mu_i)/\beta\big)}
     {\int_0^1 G_q\!\big((t-\mu_i)/\beta\big)\,dt},
\qquad \lambda\in[0,1],\;\;\beta>0,
\label{eq:lambda_density}
\end{equation}
and the synthetic instance is the convex combination
\begin{equation}
x_{\text{syn}} = (1-\lambda)\,n_{i,a}^{\text{min}} + \lambda\,n_{i,b}^{\text{min}}.
\label{eq:synthesis}
\end{equation}
Here $\beta$ is dimensionless (a fraction of the segment length) so its
interpretation is invariant across datasets and anchors, and $q$ is shared
with Phase~II.

\begin{lemma}[Properties of the placement distribution]\label{lem:placement}
For $q>1$, $\beta>0$, and any $\mu_i\in[0,1]$:
\emph{(i)} $f_i$ is a well-defined, strictly positive density on $[0,1]$, and
$x_{\text{syn}}\in\operatorname{conv}\{n_{i,a}^{\text{min}},n_{i,b}^{\text{min}}\}
\subseteq\operatorname{conv}\big(\mathcal{N}_k(i)\big)$ almost surely;
\emph{(ii)} $f_i$ is unimodal with mode at $\mu_i$;
\emph{(iii)} as $\beta\to 0^{+}$, $f_i$ converges weakly to the point mass at
$\mu_i$ (deterministic, anchor-guided placement), and as $\beta\to\infty$,
$f_i$ converges to the uniform density on $[0,1]$ (SMOTE-like placement along
the segment);
\emph{(iv)} for fixed $\beta$, larger $q$ places strictly more mass far from
$\mu_i$.
\end{lemma}
\begin{proof}
\emph{(i)} $G_q$ is continuous and strictly positive on the compact interval,
so the normalizer is finite and positive; Eq.~\eqref{eq:synthesis} with
$\lambda\in[0,1]$ is a convex combination of two elements of
$\mathcal{N}_k(i)$. \emph{(ii)} $G_q$ is strictly decreasing in $|t|$, so
$f_i$ is maximized at $\lambda=\mu_i$. \emph{(iii)} As $\beta\to0$ the mass of
$G_q((\cdot-\mu_i)/\beta)$ concentrates on any neighborhood of $\mu_i$; as
$\beta\to\infty$, $G_q((\lambda-\mu_i)/\beta)\to G_q(0)=1$ uniformly on
$[0,1]$, giving the uniform limit. \emph{(iv)} follows from the tail
asymptotics $G_q(t)\sim\big((q-1)t^2\big)^{1/(1-q)}$.
\end{proof}

Property~(iii) situates $\beta$ on an interpretable continuum between
deterministic anchor-guided placement and uniform segment coverage. Complementing this, 
property~(iv) makes $q$ a control on the propensity for excursions away from
the anchor-preferred region. The distance-dependent, heavy-tailed character of
placement is thus a property of a genuine sampling distribution, not of a
deterministic map. To implement this operational behaviour in practice, we sample from Eq.~\eqref{eq:lambda_density} by rejection sampling. Specifically, since $G_q\le 1$, propose $\lambda\sim\mathcal{U}[0,1]$ and accept with
probability $G_q((\lambda-\mu_i)/\beta)$, an exact scheme whose acceptance
rate is bounded below by $G_q(1/\beta)>0$ uniformly over anchors.

\subsection{Guarantees}\label{subsec:guarantees}

The three phases compose into two end-to-end guarantees. A per-anchor
worst-case bound that holds regardless of the random draws, and a monotone
control of that bound by the Phase~I parameter. Having established these properties qualitatively in Sect.~\ref{sec:method}; we now formalize their statements.

\begin{lemma}[Certified majority clearance]\label{lem:margin}
Every synthetic instance $x_{\text{syn}}$ seeded by anchor $x_i^{\text{min}}$
satisfies, for all $x^{\text{maj}}\in D_{\text{maj}}$,

\begin{equation}
\ell\big(x_{\text{syn}}, x^{\text{maj}}\big)\;\ge\;c_i - r_i \;=:\;\gamma_i .
\label{eq:margin}
\end{equation}
\end{lemma}

\begin{proof}
By Lemma~\ref{lem:placement}(i),
$x_{\text{syn}}\in\operatorname{conv}\big(\mathcal{N}_k(i)\big)$. Every element
of $\mathcal{N}_k(i)$ lies within distance $r_i$ of the anchor
(Eq.~\eqref{eq:extremal}), and a convex combination of points in a ball of
radius $r_i$ about $x_i^{\text{min}}$ remains in that ball, so
$\ell(x_{\text{syn}},x_i^{\text{min}})\le r_i$. The triangle inequality then
gives, for any $x^{\text{maj}}$,
\[
\ell(x_{\text{syn}},x^{\text{maj}})
\;\ge\;\ell(x_i^{\text{min}},x^{\text{maj}}) - \ell(x_{\text{syn}},x_i^{\text{min}})
\;\ge\; c_i - r_i,
\]
since $\ell(x_i^{\text{min}},x^{\text{maj}})\ge c_i$ by the definition of
$c_i$ as the distance to the nearest majority instance.
\end{proof}

We call $\gamma_i$ the \emph{certificate} of anchor
$x_i^{\text{min}}$, and an anchor \emph{certified} when $\gamma_i>0$. For a
certified anchor, no synthetic instance it seeds can approach the majority
class closer than $\gamma_i$, irrespective of which neighbors Phase~II selects
or where Phase~III places the coefficient. Anchors with $\gamma_i\le 0$ are not certified. Typically members of small minority clusters whose $k$-th neighbor lies across
a class boundary, these instances rely on the locality kernel of
Phase~II (Lemma~\ref{lem:phase2}(ii)) to polynomially suppresses the offending
cross-cluster candidates, though without a positive guarantee.


The \emph{certified fraction} $\lvert\{i:\gamma_i>0\}\rvert / N_1$
measures what proportion of the minority class admits a distance
guarantee at the chosen $k$. A high value indicates a minority class
whose local neighborhoods sit clear of the majority class; a low value
indicates one whose neighborhoods bridge into majority territory, so
that most synthesis proceeds without a positive guarantee. It is
computable at fit time, before any classifier is trained, and is
reported throughout in Sect.~\ref{sec:results}.

The certificate depends on the anchor only through $c_i$ and $r_i$, both fixed
by the data. The Phase~I parameter $\alpha$ does not change any individual
$\gamma_i$. Instead, it changes \emph{which} anchors are drawn, thereby shifting the
distribution of certificates among the synthetic instances actually produced.

\begin{proposition}[Certified clearance is monotone in $\alpha$]\label{prop:monotone}
Let $\gamma=\gamma_I$ with $I\sim P_\alpha$. If the safety score is comonotone
with the certificate across anchors, particularly when $s_i$ is taken as
the min--max normalization $\hat{\gamma}_i$ of $\gamma_i$, then
$\mathbb{E}_{P_\alpha}[\gamma]$ is nondecreasing in $\alpha$.
\end{proposition}

\begin{proof}
By Lemma~\ref{lem:palpha}(ii), $\alpha'>\alpha$ implies that $s_I$ under
$P_{\alpha'}$ first-order stochastically dominates $s_I$ under $P_{\alpha}$.
The expectation of any nondecreasing function of $s$ is therefore nondecreasing
in $\alpha$; when $\gamma$ is comonotone with $s$ (in particular when
$s=\hat{\gamma}$, so $\gamma$ is itself a nondecreasing function of $s$), the
claim follows.
\end{proof}

\begin{remark}[Practical score versus exact variant]\label{rem:instantiation}
CISO instantiates $s_i$ by Eq.~\eqref{eq:safety} rather than by
$\hat{\gamma}_i$. The product form separates the two interpretable risk channels, namely proximity and regional density, and admits the density-estimation reading of Sect.~\ref{subsec:foundation}. However, this architectural choice comes at the cost of making Proposition~\ref{prop:monotone} conditional on the empirical comonotonicity of $s_i$ and $\gamma_i$, a condition we subsequently verify in Sect.~\ref{sec:results}. The exact variant $s_i=\hat{\gamma}_i$, for which the proposition holds unconditionally, is retained as an ablation.
\end{remark}

\subsection{Algorithm and complexity}\label{subsec:algorithm}

Algorithm~\ref{alg:ciso} assembles the three phases. Its structure reflects a
strict separation between a \emph{build-once} stage, in which all
data-dependent structure is precomputed, and a \emph{per-sample} stage, in
which each synthetic instance is produced by three draws and one convex
combination. The per-sample loop performs no geometric recomputations, operating entirely on the pre-processed artifacts from the build stage. This architectural separation directly mirrors the \texttt{fit} / \texttt{generate} design of our implementation.

\begin{algorithm}[t]
\caption{Certified Interpolation Safe Oversampling (CISO)}
\label{alg:ciso}
\begin{algorithmic}[1]
\Require dataset $D$; neighbors $k$; temperature $\alpha$; floor
$\varepsilon_0$; tail index $q>1$; placement scale $\beta$
\Ensure balanced dataset $D'$
\Statex \textbf{Build once:}
\State partition $D$ into $D_{\text{min}},D_{\text{maj}}$;\;
       $\Delta N \gets N_0-N_1$;\; $k\gets\min(k,\,N_1-1,\,N_0)$
\State build $\mathcal{M}_k(i),\mathcal{N}_k(i)$ and the extremal statistics
       $c_i,r_i$ \Comment{Eqs.~\eqref{eq:maj_neighborhood}--\eqref{eq:extremal}}
\State compute the majority density field $\hat{\rho}$ and clearance field
       $\hat{c}$ \Comment{Eq.~\eqref{eq:density_field}}
\State compute $\rho_{\text{Reg}}(i)$, safety scores $s_i$, and the
       distribution $P_\alpha$ \Comment{Eqs.~\eqref{eq:rhoreg}--\eqref{eq:anchor_dist}}
\State compute scales $h_i$ and weight vectors $\tilde{w}_{i\cdot}$ for all $i$
       \Comment{Eqs.~\eqref{eq:medscale}--\eqref{eq:phase2_weight}}
\Statex \textbf{Per sample:}
\State $S_{\text{syn}}\gets\emptyset$
\While{$\lvert S_{\text{syn}}\rvert < \Delta N$}
  \State draw anchor $I\sim P_\alpha$ \Comment{Phase~I: where}
  \State draw $a\sim\operatorname{Cat}(\tilde{w}_{I\cdot})$,\;
         $b\sim\operatorname{Cat}(\tilde{w}_{I\cdot}\mid j\neq a)$
         \Comment{Phase~II: with whom}
  \State $\mu_I\gets$ center of $(n_{I,a}^{\text{min}},n_{I,b}^{\text{min}})$
         \Comment{Eq.~\eqref{eq:center}}
  \Repeat \Comment{Phase~III: how far}
    \State $\lambda\sim\mathcal{U}[0,1]$; accept w.p.
           $G_q\!\big((\lambda-\mu_I)/\beta\big)$
  \Until{accepted}
  \State $x_{\text{syn}}\gets(1-\lambda)\,n_{I,a}^{\text{min}}+\lambda\,n_{I,b}^{\text{min}}$
         \Comment{Eq.~\eqref{eq:synthesis}}
  \State $S_{\text{syn}}\gets S_{\text{syn}}\cup\{(x_{\text{syn}},1)\}$
\EndWhile
\State \Return $D'\gets D\cup S_{\text{syn}}$
\end{algorithmic}
\end{algorithm}

\paragraph{Complexity.}
The build stage is dominated by the $k$-NN queries. A naive implementation
computes the pairwise distances within each class, costing
$O\big((N_0^2+N_1^2)\,d\big)$ time and $O(N_0^2)$ memory for the majority
block. While this overhead is acceptable for the small benchmark datasets but not at scale. To address this, We use exact tree-based nearest-neighbor queries when the class size exceeds a threshold. This optimization reduces the neighborhood construction to an expected time complexity of $O\big((N_0+N_1)\,k\,d\log N_0\big)$ while requiring only $O\big((N_0+N_1)\,k\big)$ memory. The remaining build steps (fields, scores, $P_\alpha$, and the weight vectors) are all linear in $N_0k$ or $N_1k$. For each synthetic instance, the per-sample stage performs one categorical draw over $N_1$ anchors, two
over $k$ neighbors, and an expected $1/G_q(1/\beta)$ rejection trials to sample the
coefficient. This yields an amortized complexity of \ $O(N_1 + k)$, which remains independent of $d$ prior to the final convex combination. Consequently, the total runtime is dominated by the one-time neighbor construction; as empirically verified in Sect.~\ref{sec:results}, CISCO scales to datasets with $N=20{,}000$ and $d=166$ without difficulty.

\section{Experimental Protocol}\label{sec:protocol}

The evaluation strictly adheres to a protocol fixed prior to the experimental run. Specifically, the datasets, baselines, classifiers, metrics, hyperparameter selection, and statistical tests were all specified before any evaluation-set results. This preregistration serves as a deliberate safeguard against the degrees of freedom such as selective tuning, metric selection, post hoc test choice, which often inflate apparent gains in oversampling studies. Consequently, we report every preregistered outcome, including those that do not favor the proposed method (Sect.~\ref{sec:results}).

\subsection{Datasets}\label{subsec:datasets}

\begin{table}[!htbp]
\centering
\caption{Complete evaluation suite: 37 KEEL datasets (top) and 8 large-scale UCI datasets (bottom). Multi-class UCI datasets are binarized using the smallest-class-versus-rest rule.}
\label{tab:all_datasets}
\footnotesize

\begin{tabular*}{\linewidth}{@{\extracolsep{\fill}}lrrrr@{\hspace{12pt}}lrrrr@{}}
\toprule
\multicolumn{10}{c}{\textbf{KEEL Benchmark Suite}} \\
\midrule
Dataset & $N$ & $d$ & $N_1$ & IR & Dataset & $N$ & $d$ & $N_1$ & IR\\
\midrule
abalone9-18 & 731 & 8 & 42 & 16.4 & vehicle0 & 846 & 18 & 199 & 3.2 \\
dermatology-6 & 358 & 34 & 20 & 16.9 & vehicle1 & 846 & 18 & 217 & 2.9 \\
ecoli-0\_vs\_1 & 220 & 7 & 77 & 1.9 & vehicle2 & 846 & 18 & 218 & 2.9 \\
ecoli1 & 336 & 7 & 77 & 3.4 & vehicle3 & 846 & 18 & 212 & 3.0 \\
ecoli4 & 336 & 7 & 20 & 15.8 & vowel0 & 988 & 13 & 90 & 10.0 \\
flare-F & 1066 & 11 & 43 & 23.8 & car-good & 1728 & 6 & 69 & 24.0 \\
glass-0-1-2-3\_vs\_4-5-6 & 214 & 9 & 51 & 3.2 & winequality-red-4 & 1599 & 11 & 53 & 29.2 \\
glass-0-1-6\_vs\_2 & 192 & 9 & 17 & 10.3 & wisconsin & 683 & 9 & 239 & 1.9 \\
glass-0-1-6\_vs\_5 & 184 & 9 & 9 & 19.4 & yeast-0-5-6-7-9\_vs\_4 & 528 & 8 & 51 & 9.3 \\
glass1 & 214 & 9 & 76 & 1.8 & yeast-1-2-8-9\_vs\_7 & 947 & 8 & 30 & 30.6 \\
glass2 & 214 & 9 & 17 & 11.6 & yeast-1-4-5-8\_vs\_7 & 693 & 8 & 30 & 22.1 \\
glass4 & 214 & 9 & 13 & 15.5 & yeast-1\_vs\_7 & 459 & 7 & 30 & 14.3 \\
glass5 & 214 & 9 & 9 & 22.8 & yeast-2\_vs\_4 & 514 & 8 & 51 & 9.1 \\
glass6 & 214 & 9 & 29 & 6.4 & yeast-2\_vs\_8 & 482 & 8 & 20 & 23.1 \\
new-thyroid1 & 215 & 5 & 35 & 5.1 & yeast1 & 1484 & 8 & 429 & 2.5 \\
newthyroid2 & 215 & 5 & 35 & 5.1 & yeast4 & 1484 & 8 & 51 & 28.1 \\
pima & 768 & 8 & 268 & 1.9 & yeast5 & 1484 & 8 & 44 & 32.7 \\
segment0 & 2308 & 19 & 329 & 6.0 & yeast6 & 1484 & 8 & 35 & 41.4 \\
shuttle-c0-vs-c4 & 1829 & 9 & 123 & 13.9 &  & & & &  \\
\bottomrule
\end{tabular*}

\vspace{0.4cm} 

\begin{tabular*}{\linewidth}{@{\extracolsep{\fill}}lrrlr@{}}
\toprule
\multicolumn{5}{c}{\textbf{UCI Large-Scale Suite}} \\
\midrule
Dataset & $N$ & $d$ & Positive Class & IR\\
\midrule
Spambase & 4601 & 57 & spam (native) & 1.5 \\
Waveform & 5000 & 21 & smallest class & 2.0 \\
Statlog (Landsat) & 6435 & 36 & smallest class & 9.3 \\
Musk (v2) & 6598 & 166 & musk (native) & 5.5 \\
Page Blocks & 5473 & 10 & smallest class & 46.59 \\
Online Shoppers & 12330 & 17 & Revenue (native) & 5.5 \\
HTRU2 & 17898 & 8 & pulsar (native) & 9.9 \\
Letter & 20000 & 16 & smallest class & 26.2 \\
\bottomrule
\end{tabular*}
\end{table}

We evaluate on 50 datasets in total, partitioned into three distinct suites. The primary \emph{evaluation suite} comprises 37 binary imbalanced datasets from the KEEL repository \citep{EC8}, with imbalance ratios from $1.8$ to $41.4$ and dimensions from $3$ to $34$ (Table~\ref{tab:all_datasets}). A disjoint \emph{tuning suite} of 5 additional KEEL datasets is held out exclusively for hyperparameter selection and is strictly excluded from all evaluation tables and tests (Sect.~\ref{subsec:tuning}). A \emph{scale suite} of 8 larger datasets from the UCI repository~\citep{ci25}, with sample sizes up to $20{,}000$ and dimensions up to $166$ (Table~\ref{tab:all_datasets}), probes generalization beyond the KEEL regime. The scale suites together provide 45 datasets on which results are reported. Multi-class datasets are binarized by the smallest-class-versus-rest rule, with the target class recorded explicitly per dataset. Within each cross-validation fold, features are standardized on the training partition and the transformation applied to the test partition. Nominal attributes are one-hot encoded and missing numeric values imputed using training-fold means.

\subsection{Baseline methods}\label{subsec:baselines}
The comparison analysis evaluates three distinct tiers of baselines. \emph{Tier~1 (Baseline Controls)} comprises no resampled training, cost-sensitive class weighting, and random oversampling. These methods establish whether complex resampling is justified over standard baselines. \emph{Tier~2 (Interpolation family)} includes foundational synthetic oversampling methods: SMOTE, Borderline-SMOTE, ADASYN, SMOTE-ENN~\citep{ci27}, SMOTE-Tomek~\citep{ci28}, and KMeans-SMOTE~\citep{ci29}. \emph{Tier~3 (Density- and weight-aware relatives)}: evaluates methods designed to account for local data structure: Safe-Level-SMOTE and G-SMOTE\citep{ci30}.

All baselines methods utilize established library implementations~\citep{ci31} with default configurations. To ensure a fair comparison, all $k$-type neighbor parameters set to match CISO's neighborhood size $k$. Following our preregistered protocol, no per-dataset hyperparameters tuning is applied to any baseline, including CISO (Sect.~\ref{subsec:tuning}). Execution failures ( such as a sampler throwing an exception due to insufficient minority instances in a fold) are explicitly recorded rather than handled. When a method fails on a given fold, it falls back to no resampling for that fold only, and the failure is logged and reported in Sect.~\ref{sec:results}.


\subsection{Classifiers}\label{subsec:classifiers}
Each resampled training set is evaluated with four classifiers. These represent distinct inductive biases, including logistic regression (LR), an RBF-kernel support vector machine (SVC), random forest (RF), and histogram-based gradient boosting (HGB). Results are analyzed separately for each classifier rather than pooled. Since resampling techniques interact differently across classifier families, aggregation would obscure these distinct behavioral interactions~\citep{c34}. For the large-scale suit, the support vector machine is excluded due to computational constraints, as preregistered.

\subsection{Evaluation metrics}\label{subsec:metrics}
The primary metric is the area under the precision--recall curve (PR-AUC, equivalently average precision). This threshold-free metric is well suited for imbalance regimes, where precision--recall analysis offers greater informativeness than the ROC curve~\citep{ci32}. We additionally report two secondary metrics, namely the area under the ROC curve (ROC-AUC) and the geometric mean~\citep{EC2C} of class-wise recalls,
\begin{equation}
\text{G-mean} = \sqrt{\text{TPR}\cdot\text{TNR}},
\label{eq:gmean}
\end{equation}
evaluated at the standard decision threshold.

For the support vector machine, ranking metrics are calculated using the raw decision function, while G-mean uses its sign. For probabilistic classifiers, ranking metrics rely on the predicted positive-class probability, whereas G-mean uses a threshold of $0.5$. In addition to predictive metrics, we report the fit-time diagnostics introduced in Sect.~\ref{sec:method}, specifically the certified fraction and the held out maximum mean  discrepancy between synthetic and unseen minority instances.

\subsection{Hyperparameter protocol}\label{subsec:tuning}
CISO has five hyperparameters, including the neighborhood size $k$, temperature $\alpha$, floor $\varepsilon_0$, tail index $q$, and placement scale $\beta$. These are selected \emph{once} using only the 5 tuning datasets listed in Table~\ref{tab:tuning} via grid search over $k\in\{5,7\}$, $\alpha\in\{-1,0,0.5,1,2\}$, $q\in\{1.2,1.5,2.0\}$, $\beta\in\{0.1,0.2,0.4\}$, with $\varepsilon_0=0.1$ fixed. This process yields $90$ total configurations. The selection criterion, fixed prior to evaluation, is the mean PR-AUC across the tuning datasets and the four classifiers. Ties are resolved first by higher G-mean and subsequently by smaller absolute temperature $\lvert\alpha\rvert$.  

Once selected, the optimal configuration is frozen, and the evaluation suite is run exactly once using the frozen setup. Crucially, the tuning and evaluation suites remain strictly disjoint. Furthermore, no baseline receives per-dataset tuning, ensuring every method is evaluated under a single global configuration. This symmetrical protocol eliminates tuning bias that would otherwise favor the proposed method.

\begin{table}[!htbp]
\centering
\caption{Tuning suite: the 5 KEEL datasets used exclusively for hyperparameter selection, excluded from all evaluation tables and tests.}
\label{tab:tuning}
\label{tab:tuning}
\footnotesize
\begin{tabular*}{0.65\linewidth}{@{\extracolsep{\fill}}lrrr@{}}
\toprule
Dataset & $N$ & $d$ & IR \\
\midrule
glass0    & 214  & 9 & 2.1 \\
haberman  & 306  & 3 & 2.8 \\
ecoli3    & 336  & 7 & 8.6 \\
yeast3    & 1484 & 8 & 8.1 \\
abalone19 & 4174 & 8 & 129.4 \\
\bottomrule
\end{tabular*}
\end{table}

\subsection{Statistical analysis}\label{subsec:stats}
For each combination of classifier and metric, we compute each method's mean rank across
the evaluation datasets. We then apply the Friedman test to evaluate the global null hypothesis that all methods perform equally. When the null hypothesis is rejected, we perform Holm-corrected post hoc comparisons of CISO against every other method
\citep{ci33}. This procedure controls the family-wise error rate over control-versus-all comparisons rather than all pairwise combinations.

Effect sizes are reported as win, tie and loss counts, where a tie is defined as $\lvert\Delta\rvert<10^{-3}$. Pairwise comparisons of CISO against each competing method use the Wilcoxon signed-rank test on per-dataset scores, with Holm correction applied across the family of control-versus-all comparisons. Ablation comparisons (Sect.~\ref{subsec:ablation}) use the same test with Holm correction applied across the ablation family. The certificate-as-predictor is tested by Spearman rank correlation between the certified fraction and the per-dataset PR-AUC advantage, computed per classifier.

\subsection{Preregistered questions and decision rules}\label{subsec:questions}

The experimental protocol defined four \emph{confirmatory} research questions, each accompanied by an explicit decision rule. In addition, four \emph{descriptive} analyses were formulated without pass or fail criteria. All research questions and analytical procedures were established strictly prior to executing the evaluation suite.

\paragraph{Confirmatory questions.}
\begin{itemize}
\item[(Q1)] \emph{Predictive performance.} Does CISO outperform the baseline tiers on the primary metric? \textbf{Rule:} Success requires a superior mean PR-AUC rank compared to every Tier-2 method across at least three of the four classifiers, together with a Holm-adjusted $p<0.05$ against SMOTE. Failure to outperform Tier-1 (class weighting, random oversampling) on any classifier is to be reported prominently.
\item[(Q2)] \emph{Component contribution.} Does each phase contribute
measurably? \textbf{Rule:} For each ablation, the full method is compared against the ablated variant across all evaluation datasets using a Wilcoxon signed-rank test. The resulting p-values are adjusted using Holm correction across the ablation family. A component is confirmed to be effective only if the test yields a statistically significant result ($p<0.05$) in the expected direction. 

\item[(Q3)] \emph{Certificate as predictor.} Does the certified fraction
predict where CISO gains over SMOTE? \textbf{Rule:} Evaluation relies on the Spearman correlation between the per-dataset certified fraction and the per-dataset PR-AUC difference (CISO minus SMOTE). The preregistered hypothesis posits a \emph{negative} correlation, i.e.\ larger advantage on datasets with lower certification.
\item[(Q4)] \emph{Distributional fidelity.} Are CISO's synthetic instances
closer in distribution to held-out minority data than those of competing
samplers? \textbf{Rule:} Success requires significantly lower held-out
$\mathrm{MMD}^2$ than at least three of the four comparison samplers
(Wilcoxon, Holm-corrected) and no significantly higher value against any.
Folds with fewer than three held-out minority instances are excluded,
as $\mathrm{MMD}^2$ is not estimable; datasets retaining fewer than eight usable folds are excluded entirely.
\end{itemize}

\paragraph{Descriptive analyses.}
We additionally pre-specified four descriptive analyses that do not involve explicit pass or fail criteria. These comprise parameter sensitivity across the tuning grid, wall-clock computational cost for both the build-once and per-sample stages, sampler reliability, measured as the fraction of folds on which each method fails, and generalization performance on the scale suite.


\section{Results}\label{sec:results}

We report the preregistered analyses in the sequence established in Sect.~\ref{sec:protocol}, evaluating each confirmatory question against its predefined decision rule. Following our pre-specified protocol, methods that experience severe execution failures on many of datasets are excluded from the primary predictive comparisons and evaluated in detail within the reliability analysis (Sect.~\ref{subsec:reliability}). Therefore, all primary evaluations compare CISO against the remaining baseline models across the full suite of benchmark datasets.

\subsection{Predictive performance}\label{subsec:rq1}

\begin{figure}
    \centering
    \includegraphics[width=\textwidth]{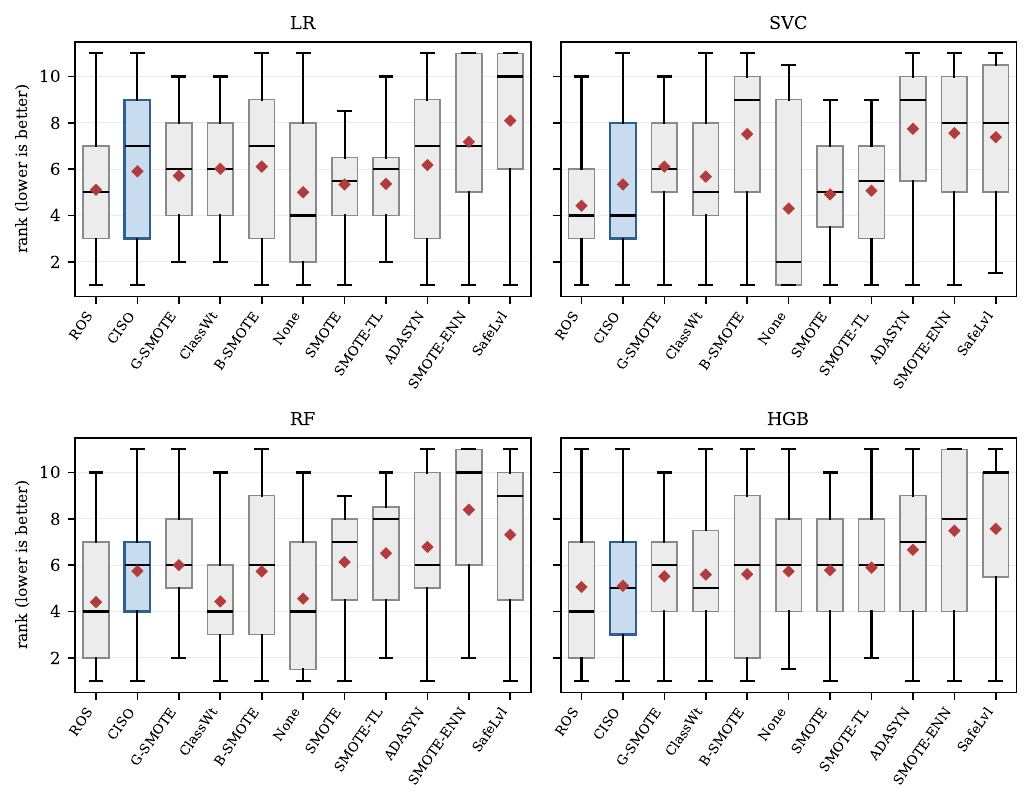}
    \caption{Distribution of per-dataset PR-AUC ranks across the 37 evaluation
  datasets, by classifier (lower is better). Boxes show the interquartile
  range and median; red diamonds mark mean ranks; CISO is highlighted.
  Methods are ordered by mean rank under HGB. The extensive overlap of the distributions is the substance of the equivalence result in Table~\ref{tab:rq1}.}
    \label{fig:rankbox}
\end{figure}

Table~\ref{tab:rq1} summarizes mean ranks, mean PR-AUC, win, tie and loss counts, along with Holm-corrected $p$-values for CISO against each competing method across all four classifiers. Fig.~\ref{fig:rankbox} illustrates the complete distribution of per-dataset ranks. The Friedman test rejects the global null for all four classifiers ($p \le 3.6\times10^{-3}$), confirming that performance across the evaluated methods is not interchangeable in aggregate. However, a detailed analysis of these rank distributions reveals a more nuanced pattern than uniform superiority for any single method.

\begin{table}[t]\centering
\caption{Q1: PR-AUC on the 37-dataset evaluation suite. For each classifier: mean rank across datasets (lower is better), mean PR-AUC, win/tie/loss of CISO against the method, and the Holm-corrected $p$-value of the Wilcoxon signed-rank test of CISO against it. Friedman $p$-values: LR~$=$~0.0006, SVC~$=$~8.8e-10, RF~$=$~1.9e-08, HGB~$=$~0.0036.}
\label{tab:rq1}\footnotesize\setlength{\tabcolsep}{3.5pt}
\begin{tabular}{@{}lrrcc rrcc@{}}\toprule
 & \multicolumn{4}{c}{LR} & \multicolumn{4}{c}{SVC}\\
\cmidrule(lr){2-5}\cmidrule(lr){6-9}
Method & rank & PR-AUC & W/T/L & Holm & rank & PR-AUC & W/T/L & Holm\\\midrule
No resampling & 5.00 & 0.693 & 13/8/16 & 0.993 & 4.30 & 0.727 & 10/6/21 & 0.549 \\
Class weighting & 6.01 & 0.693 & 16/5/16 & 1.000 & 5.68 & 0.710 & 15/9/13 & 1.000 \\
Random oversampling & 5.11 & 0.699 & 16/4/17 & 1.000 & 4.42 & 0.718 & 16/4/17 & 1.000 \\
SMOTE & 5.34 & 0.702 & 11/10/16 & 1.000 & 4.91 & 0.711 & 15/7/15 & 0.866 \\
Borderline-SMOTE & 6.11 & 0.695 & 17/3/17 & 1.000 & 7.51 & 0.699 & 22/5/10 & 0.205 \\
ADASYN & 6.18 & 0.697 & 17/4/16 & 1.000 & 7.74 & 0.697 & 19/5/13 & 1.000 \\
SMOTE-ENN & 7.18 & 0.689 & 21/5/11 & 0.919 & 7.55 & 0.692 & 26/4/7 & 0.071 \\
SMOTE-Tomek & 5.36 & 0.702 & 13/6/18 & 1.000 & 5.07 & 0.710 & 16/6/15 & 1.000 \\
Safe-Level-SMOTE & 8.09 & 0.669 & 24/4/9 & 0.006 & 7.38 & 0.687 & 24/6/7 & 0.001 \\
G-SMOTE & 5.72 & 0.699 & 13/6/18 & 1.000 & 6.11 & 0.707 & 16/7/14 & 1.000 \\
\textbf{CISO} & 5.91 & 0.694 & -- & -- & 5.34 & 0.715 & -- & -- \\
\midrule
 & \multicolumn{4}{c}{RF} & \multicolumn{4}{c}{HGB}\\
\cmidrule(lr){2-5}\cmidrule(lr){6-9}
Method & rank & PR-AUC & W/T/L & Holm & rank & PR-AUC & W/T/L & Holm\\\midrule
No resampling & 4.55 & 0.739 & 11/5/21 & 0.411 & 5.73 & 0.716 & 17/8/12 & 1.000 \\
Class weighting & 4.43 & 0.736 & 9/7/21 & 0.453 & 5.59 & 0.716 & 17/5/15 & 0.775 \\
Random oversampling & 4.41 & 0.736 & 9/7/21 & 0.382 & 5.05 & 0.731 & 18/8/11 & 1.000 \\
SMOTE & 6.14 & 0.726 & 17/10/10 & 0.416 & 5.78 & 0.731 & 15/10/12 & 1.000 \\
Borderline-SMOTE & 5.73 & 0.733 & 17/5/15 & 0.819 & 5.61 & 0.734 & 17/4/16 & 0.925 \\
ADASYN & 6.78 & 0.720 & 19/9/9 & 0.440 & 6.66 & 0.723 & 19/7/11 & 0.552 \\
SMOTE-ENN & 8.39 & 0.703 & 24/7/6 & 0.002 & 7.49 & 0.702 & 24/2/11 & 0.115 \\
SMOTE-Tomek & 6.51 & 0.725 & 20/9/8 & 0.440 & 5.89 & 0.730 & 18/7/12 & 1.000 \\
Safe-Level-SMOTE & 7.31 & 0.718 & 19/5/13 & 0.433 & 7.57 & 0.713 & 27/1/9 & 0.009 \\
G-SMOTE & 6.00 & 0.722 & 17/11/9 & 0.549 & 5.51 & 0.729 & 18/8/11 & 0.641 \\
\textbf{CISO} & 5.74 & 0.729 & -- & -- & 5.11 & 0.733 & -- & -- \\
\bottomrule\end{tabular}\end{table}

Across all four classifiers, CISO demonstrates performance that is statistically indistinguishable from SMOTE. The Holm-corrected Wilcoxon tests yield adjusted $p$-values of $p = 1.000$ (LR), $0.866$ (SVC), $0.416$ (RF) and $1.000$ (HGB). Correspondingly, win/tie/loss counts remains close to even, showing $15$ wins, $7$ ties, and $15$ losses under SVC, and $17$ wins, $10$ ties, and $10$ losses under RF. Across every classifier evaluated, the difference in mean PR-AUC between CISO and SMOTE stays consistently below $0.005$. A similar pattern holds when compared against the Tier-1 baselines. Therefore, no comparison
against class weighting, random oversampling, or unresampled data reaches statistical significance under any classifier.

CISO achieves its strongest placement under HGB, ranking second ($5.11$) behind random oversampling ($5.05$) among eleven methods. Under SVC it ranks fourth ($5.34$), while occupying intermediate ranks
under LR ($5.91$) and RF ($5.74$), where unresampled or Tier-1 approaches yield the top mean ranks. The comparisons that remain statistically significant after Holm correction are those that CISO wins. Specifically, CISO significantly outperforms Safe-Level-SMOTE across three classifiers ($p = 0.006$ for LR, $0.001$ for SVC, $0.009$ for HGB) and SMOTE-ENN under RF ($p = 0.002$).

Two principal conclusions follow from these evaluations. First, the preregistered success criterion (Q1) is not met. CISO does not achieve a superior mean rank over every Tier-2 method on three or more classifiers, nor does it reach significance over SMOTE on any classifier. Second, no sampling method in the comparison, regardless of algorithmic complexity, consistently outperforms the Tier-1 baselines on the primary threshold-free metric. As Fig.~\ref{fig:rankbox} shows, the per-dataset rank distributions of the leading methods overlap substantially, such that their mean-rank ordering reflects a spread too broad to separate them statistically. Under a preregistered protocol spanning four classifiers and Holm-corrected testing, this second point aligns with conclusions reached by recent systematic benchmarks~\citep{ci14,ci15}.

Taken together, these results place CISO at parity with the strongest members of its algorithmic class, ranking second among eleven under HGB and significantly ahead of two established baselines. At the same time, the findings demonstrate that predictive headroom available to geometric refinement of
interpolation oversampling on threshold-free metrics with modern classifiers is limited for every method tested, not only CISO. Statistical equivalence rather than dominance is therefore the realistic outcome under this evaluation protocol. Consequently, the formal guarantees established
in Sect.~\ref{sec:method} and the reliability demonstrated in Sect.~\ref{subsec:reliability} are evaluated on their own terms in what follows, alongside this competitive predictive result rather than
in place of it.

\subsection{Component ablation}\label{subsec:ablation}

To evaluate whether each phase contributes measurably, we compare the full CISO method against four variants, each systematically disabling or replacing exactly one component while holding all other parameters fixed. Phase~I anchor selection is disabled by setting $\alpha=0$, which reduces anchor selection to the uniform draw defined in Lemma~\ref{lem:palpha}(i). The Phase~II clearance factor is removed from the weight calculation, leaving only the locality kernel. Phase~III placement is disabled by drawing the interpolation coefficient uniformly, $\lambda\sim\mathcal{U}[0,1]$, rather than from the
$q$-Gaussian density defined in Eq.~\eqref{eq:lambda_density}. Finally, the practical safety score in Eq.~\eqref{eq:safety} is replaced by the exact certificate variant $s_i=\hat{\gamma}_i$ defined in
Remark~\ref{rem:instantiation}.

All remaining hyperparameters are held at their locked configuration across all runs. Table~\ref{tab:ablation} reports two distinct metrics: predictive performance across the four classifiers, and the geometric novelty diagnostic, which measures the mean distance from synthetic instances to the nearest original minority instance.

On predictive performance, the result is a clean null. No individual component reaches statistical significance on any classifier after Holm correction across the ablation suite. The smallest adjusted $p$-value among all sixteen comparisons is $0.059$ for Phase~I under RF. Furthermore, the median per-dataset PR-AUC differences remain below $0.0016$ in magnitude, which is three orders of magnitude smaller than the between-dataset spread of the metric itself. The preregistered criterion for Q2 is therefore not confirmed. Per the protocol, we weaken the corresponding claim without reanalysis, and we do not assert that any individual phase of CISO improves predictive performance.

On geometry the same comparisons behave oppositely. Every component is significant on the novelty diagnostic, with Holm-corrected $p$-values from $0.007$ down to below $10^{-7}$, and effect sizes two orders of magnitude larger than on the predictive axis. Specifically, median novelty differences reach $+0.294$ for Phase~III, $+0.119$ for Phase~I, $+0.091$ for the safety score, and $+0.011$ for the Phase~II clearance factor. Each component shifts synthetic instances substantially in the intended direction. For instance, disabling stochastic placement or safety-guided seeding draws synthetic instances measurably closer to existing minority samples. The certified fraction, in contrast, remains identical across all five variants at $0.244$, exactly as expected. This stability arises because the
certificate $\gamma_i = c_i - r_i$ depends strictly on the data geometry and $k$, rather than on which operational phases are active (Sect.~\ref{subsec:guarantees}).

\begin{table*}[t]
\centering
\caption{Component ablation. For each variant we report the median per-dataset PR-AUC difference (CISO minus variant) by classifier, the smallest Holm-corrected $p$-value across the four classifiers, and the corresponding comparison on the geometric novelty diagnostic (mean distance from synthetic instances to the nearest original minority instance). No component is detectable in predictive performance; every component is detectable in geometry.}
\label{tab:ablation}
\renewcommand{\arraystretch}{2.3}
\resizebox{\textwidth}{!}{%
\begin{tabular}{@{}llrrrrc rc@{}}\toprule
& & \multicolumn{5}{c}{$\Delta$PR-AUC (CISO $-$ variant)} & \multicolumn{2}{c}{$\Delta$novelty}\\
\cmidrule(lr){3-7}\cmidrule(lr){8-9}
Component removed & Variant & LR & SVC & RF & HGB & min Holm $p$ & median & Holm $p$\\\midrule
Phase I (safety-guided anchors) & $\alpha=0$ & +0.0000 & +0.0012 & +0.0012 & -0.0005 & 0.059 & +0.119 & $<$0.001 \\
Phase II clearance factor & kernel only & +0.0000 & +0.0000 & +0.0000 & +0.0006 & 0.392 & +0.011 & 0.007 \\
Phase III (q-Gaussian placement) & $\lambda\sim\mathcal{U}[0,1]$ & +0.0000 & -0.0010 & +0.0016 & -0.0001 & 0.090 & +0.294 & $<$0.001 \\
practical safety score & $s_i=\hat\gamma_i$ & -0.0000 & +0.0000 & +0.0000 & +0.0000 & 0.586 & +0.091 & $<$0.001 \\
\bottomrule
\end{tabular}%
}
\end{table*}

Interpreted together, the two halves of Table~\ref{tab:ablation} explain more than either does alone. The components are not inert, as they demonstrably govern the geometry of synthesis with high statistical confidence. However, that geometric control does not translate into predictive gain. This finding is entirely consistent with the performance ceiling established in Sect.~\ref{subsec:rq1}. When eleven distinct methods spanning naive duplication to sophisticated density weighting, are already statistically indistinguishable on PR-AUC, refinements to placement geometry have no signal margin to exploit. Consequently, We describe CISO's phases as provide formal guarantees and geometric control at zero predictive cost, rather than as performance-improving mechanisms in their own right. Under this framing, the absence of a predictive penalty serves as the primary empirical claim supported by these evaluations.

\subsection{Certificate as predictor}\label{subsec:certificate}

The certified fraction is computable at fit time, before any classifier is trained, which makes it a candidate rule for deciding in advance where CISO is likely to help. The preregistered hypothesis (Q3) was that this quantity would negatively correlate with CISO's advantage over SMOTE. The underlying expectation was that datasets whose minority class admits few distance guarantees are geometrically harder, making them the regimes where safety-guided synthesis pays off most.

The certified fraction varies widely across the evaluation suite, so the test is well posed. Its median is $0.042$ and its mean $0.244$, with 13 of the 37 datasets admitting no certified anchor at all and two exceeding $0.9$. A zero certified fraction does not prevent synthesis. Every anchor retains a positive seeding probability (Lemma~\ref{lem:palpha}(iii)) regardless of whether its certificate is positive. Consequently, CISO generates the full complement of $\Delta N$ instances on these datasets as on any other, meaning the guarantee of Lemma~\ref{lem:margin} is vacuous rather than restrictive when $\gamma_i \le 0$. Table~\ref{tab:q3} reports the Spearman rank correlation between the per-dataset certified fraction and the per-dataset PR-AUC difference between CISO and SMOTE, computed separately for each classifier. 

\begin{table}[t]
\centering
\caption{Q3: Spearman rank correlation between the per-dataset
certified fraction and the per-dataset PR-AUC advantage of CISO over
SMOTE, across the 37 evaluation datasets. The preregistered hypothesis
was a significant negative correlation.}
\label{tab:q3}
\begin{tabular*}{0.85\linewidth}{@{\extracolsep{\fill}}lrrr@{}}
\toprule
Classifier & $\rho$ & $p$ & median advantage\\
\midrule
LR  & $+0.026$ & 0.878 & $-0.0002$\\
SVC & $+0.249$ & 0.137 & $\phantom{-}0.0000$\\
RF  & $+0.042$ & 0.805 & $+0.0003$\\
HGB & $+0.011$ & 0.947 & $+0.0002$\\
\bottomrule
\end{tabular*}

\end{table}

The criterion for Q3 is \emph{not supported}. No correlation reaches significance on any classifier, the largest being $\rho=+0.249$ under SVC at $p=0.137$. Moreover, all four are weakly positive rather than negative, so the hypothesis fails on direction as well as on magnitude. The median advantage over SMOTE is within $\pm 0.0003$ on every classifier, which is consistent with the equivalence reported in Sect.~\ref{subsec:rq1}. When two methods differ this little in aggregate, there is little signal for any dataset-level property to predict.

We report this null without qualification. The certificate is a worst-case geometric statement about individual synthetic instances, proved from $c_i$ and $r_i$ alone. Whether a given dataset yields many or few such guarantees does not, on this evidence, forecast how the
resulting synthetic data will affect a downstream classifier. That outcome depends additionally on the classifier family, the decision boundary, and the interaction between synthesis and the learning algorithm. Section~\ref{subsec:disc-when} discusses the boundary this places on how the certified fraction should be read.

\subsection{Distributional fidelity}\label{subsec:fidelity}
 
The analyses in Sects.~\ref{subsec:rq1}--\ref{subsec:ablation} evaluate methods by their effect on a downstream classification performance. We now examine the synthetic instances directly to assess how closely they approximate the underlying minority distribution they augment. Standard training-set geometric measures inherently favor memorization, since a sampling procedure that merely duplicates existing instances scores
perfectly under any training-distance metric. To prevent this bias, we evaluate synthetic quality against held-out minority instances instead.

For each fold, every sampler is fitted on the training partition, and its synthetic block is evaluated against the test partition's minority instances, which remain unseen during generation. We report the unbiased squared maximum mean discrepancy ($\mathrm{MMD}^2$, using an RBF kernel with a per-fold median bandwidth) as our primary metric, supplemented by a classifier two-sample test (C2ST) as a corroborating diagnostic. Random oversampling serves as a baseline reference, since it reproduces the empirical minority distribution exactly, it establishes the noise floor for the metric. This evaluation includes all 35 datasets that satisfy the preregistered exclusion criteria. 

\begin{figure*}[t]\centering
  \includegraphics[width=\textwidth]{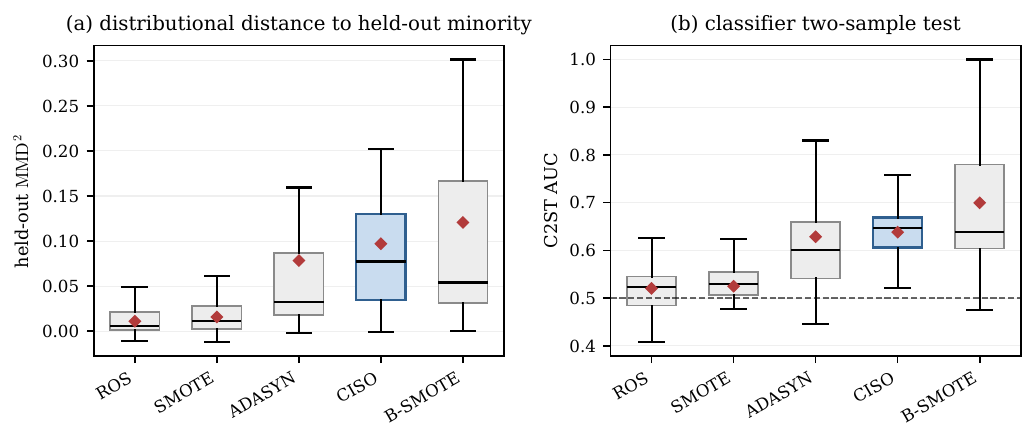}
  \caption{Distributional fidelity of synthetic instances against held-out
  minority data, over 35 datasets. (a) Unbiased $\mathrm{MMD}^2$ (lower is
  better); (b) classifier two-sample test AUC, where $0.5$ (dashed) denotes
  indistinguishability from real data. Boxes show interquartile range and
  median; red diamonds mark means. Both panels order the samplers identically,
  by the strength of the distributional reweighting each performs.}
  \label{fig:fidelity}
\end{figure*}

Figure~\ref{fig:fidelity} presents the overall performance, establishing a clear hierarchy across the evaluated methods. ROS achieves the lowest $\mathrm{MMD}^2$ at $0.0111$, followed by SMOTE ($0.0159$), ADASYN ($0.0783$), CISO ($0.0971$), and Borderline-SMOTE ($0.1206$). CISO's held-out $\mathrm{MMD}^2$ is significantly higher than that of both ROS and SMOTE, indicating lower distributional fidelity. Further, the corresponding Holm-corrected significance values are $p = 2.3\times 10^{-10}$ and $1.8\times 10^{-10}$, with median differences of $+0.058$ and $+0.051$ respectively. Conversely, CISO is statistically indistinguishable from ADASYN ($p = 0.091$) and Borderline-SMOTE ($p = 0.704$). Because the preregistered hypothesis (Q4) required significantly lower $\mathrm{MMD}^2$ than at least three of the four baseline samplers, it is not supported. The C2ST results independently mirror this relative ordering, ranging from $0.520$ for ROS (which is indistinguishable from real held-out data) to 0.700 for Borderline-SMOTE, with CISO recording $0.638$.

The ordering itself is the more informative result, because it is systematic rather than arbitrary. The five samplers are ranked exactly by how much each deliberately reweights the minority distribution. ROS applies none; SMOTE smooths locally without directional preference; ADASYN concentrates toward hard, boundary-adjacent instances; CISO, at the locked $\alpha=+2$, shifts weight toward safe, interior instances; and Borderline-SMOTE concentrates entirely on the boundary. Fidelity to the source distribution decreases monotonically along this continuum, which represents a property of the method family rather than of any single member. Consequently, any deliberate departure from uniform seeding costs fidelity in proportion to its strength.

\begin{figure}[t]
\centering
  \includegraphics[width=\linewidth]{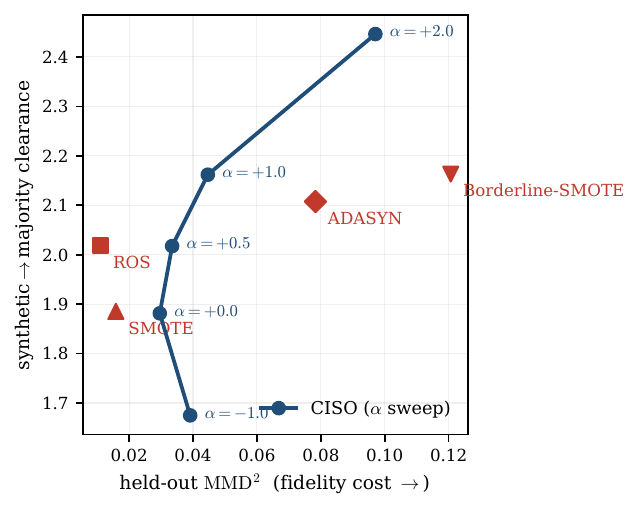}
  \caption{The fidelity--safety plane. The $\alpha$ sweep (blue) traces a
  continuous curve: clearance increases monotonically with $\alpha$, while
  held-out $\mathrm{MMD}^2$ is minimized at $\alpha=0$ and rises in both
  directions. Fixed competing samplers (red) occupy isolated points in the
  same plane. Means over 35 datasets.}
  \label{fig:frontier}
\end{figure}

That this is a mechanism rather than a coincidence is shown directly by CISO's own parametric behaviour. Sweeping $\alpha\in\{-1,0,0.5,1,2\}$ over the same 35 datasets while measuring both quantities per fold yields two clean trends. Clearance, defined as the mean
distance from synthetic instances to the nearest majority instance, is perfectly monotone in $\alpha$. As, it rises from $1.675$ at $\alpha=-1$ to $2.446$ at $\alpha=+2$, providing the empirical counterpart of Proposition~\ref{prop:monotone}. Held-out $\mathrm{MMD}^2$, by contrast, is U-shaped with its minimum exactly at $\alpha=0$ ($0.0296$), rising to $0.0391$ at $\alpha=-1$ and $0.0971$ at $\alpha=+2$. Both departures from uniform seeding, whether directed toward the boundary or toward the interior, reduce distributional fidelity. The setting $\alpha=0$ thus represents the fidelity-optimal configuration within the family, aligning exactly as the reweighting interpretation of Eq.~\eqref{eq:anchor_dist} predicts.

Figure~\ref{fig:frontier} plots both metric on the shared axes. The $\alpha$ parameter sweep traces a continuous curve through the fidelity-clearance plane, whereas each fixed competitor occupies a single point within this space. Two key observations follow from this visualization. First, the fixed baselines are not enclosed by the curve, as ROS and SMOTE attains lower $\mathrm{MMD}^2$ values at comparable clearance. Consequently, CISO does not dominate these methods within this plane. Second, the upper region of the curve reaches clearance levels, $2.16$ at $\alpha=1$ and $2.45$ at $\alpha=2$ that no fixed baseline attains. Meanwhile, ADASYN ($0.078$, $2.11$) and Borderline-SMOTE ($0.121$, $2.16$) incur comparable or greater fidelity cost for less clearance. The practical interpretation is that fidelity and clearance trade off inherently within this framework. Existing methods occupy fixed points along that trade-off, whereas a single signed parameter continuously traverses the entire spectrum without requiring a change of algorithm.

\subsection{Generalization to larger data}\label{subsec:scale}

The KEEL evaluation suite represents the standard benchmarks on which oversampling methods are conventionally assessed, but its datasets remain small, as none exceeds $N=2308$ or $d=34$. Whether the conclusions established in Sects.~\ref{subsec:rq1}--\ref{subsec:fidelity} persist at larger scale constitutes a separate empirical question. The scale suite of eight UCI datasets (Table~\ref{tab:all_datasets}) addresses this question directly, extending up to $N=20{,}000$ and $d=166$. In accordance with the preregistered protocol, the support vector machine is omitted from this experiment for computational tractability. Consequently, here results are reported across logistic regression, random forest, and gradient boosting classifiers.

\begin{table}[b]\centering
\caption{Scale suite: mean PR-AUC rank across the eight UCI datasets (lower is better). The final rows give CISO's median per-dataset PR-AUC difference against SMOTE and the Wilcoxon signed-rank $p$-value. The support vector machine is omitted per the preregistered protocol.}
\label{tab:scale-results}\footnotesize
\begin{tabular*}{0.85\linewidth}{@{\extracolsep{\fill}}lrrr@{}}
\toprule
Method & LR & RF & HGB\\\midrule
No resampling & 2.69 & 4.69 & 4.19 \\
Class weighting & 5.12 & 3.38 & 4.50 \\
Random oversampling & 5.62 & 3.00 & 3.00 \\
SMOTE & 5.50 & 4.50 & 4.88 \\
Borderline-SMOTE & 9.75 & 7.38 & 7.62 \\
ADASYN & 7.62 & 8.12 & 6.50 \\
SMOTE-ENN & 7.75 & 10.88 & 10.50 \\
SMOTE-Tomek & 5.88 & 5.50 & 5.38 \\
Safe-Level-SMOTE & 4.94 & 7.19 & 9.44 \\
G-SMOTE & 6.62 & 5.62 & 4.50 \\
\textbf{CISO} & 4.50 & 5.75 & 5.50 \\
\midrule
CISO $-$ SMOTE (median) & +0.0016 & -0.0004 & -0.0003 \\
Wilcoxon $p$ & 0.383 & 0.641 & 0.547 \\
\bottomrule\end{tabular*}
\end{table}

Table~\ref{tab:scale-results} reports the outcome. The predictive performance pattern replicates cleanly on the larger datasets. Across the three evaluated classifiers, CISO's median per-dataset PR-AUC difference against SMOTE remains within $\pm0.0016$, and no comparison against SMOTE reaches significance ($p\ge0.38$). Mean ranks place CISO second of eleven under logistic regression and mid-field under random forest and gradient boosting. Random oversampling and class weighting consistently remain among the strongest, which mirrors the pattern observed on the evaluation suite, at an order of magnitude more data and five times the dimensionality. Consequently, the performance equivalence established in Sect.~\ref{subsec:rq1} does not represent an artifact of small-sample benchmarks.

Two further findings from this suite carry greater significance than the overall ranks. First, the tree-based search path scales in accordance with theoretical predictions, with computational cost growing near-linearly to $1.23$\,s at $N=20{,}000$ (Sect.~\ref{subsec:runtime}). As a result, the $O(N^2)$ dense matrix calculation is never
required at these sample sizes. Second, CISO completed every fold of every dataset, including musk at $d=166$, the highest dimensionality in the study. In contrast, two competing baseline samplers failed to execute under these conditions, which motivates the reliability analysis that follows.

\subsection{Parameter sensitivity}\label{subsec:sensitivity}

Sensitivity is assessed strictly on the tuning suite. This restriction is required by the protocol, since the 37 evaluation datasets are reserved for the confirmatory analyses in  Sects.~\ref{subsec:rq1}--\ref{subsec:ablation}. Characterizing parameter behaviour on the evaluation suite would consume the very data the preregistration protects. Consequently, all parameter effects reported here are therefore computed over the five tuning datasets and the full 90-configuration grid detailed in Sect.~\ref{subsec:tuning}, representing $27{,}000$ fold-level evaluations in total. The findings presented in this section are descriptive rather than confirmatory.

\begin{figure*}[b]
  \centering
  \includegraphics[width=\textwidth]{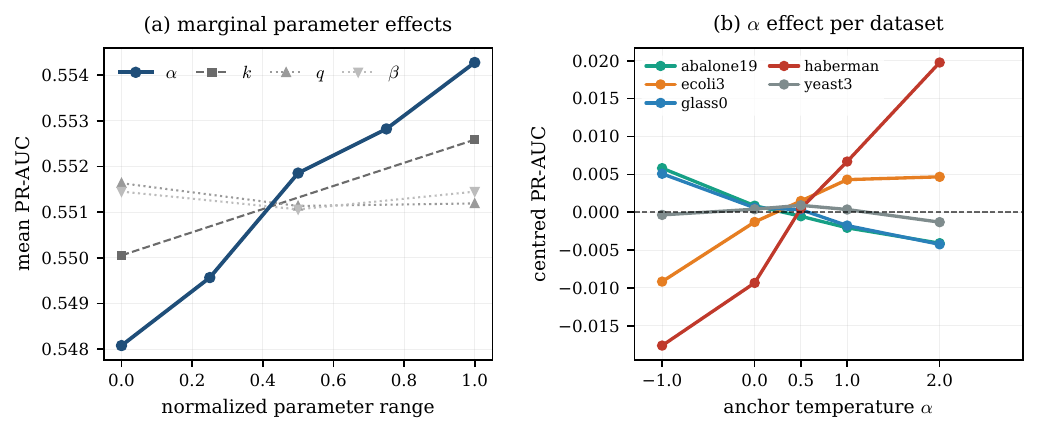}
  \caption{Hyperparameter sensitivity over the five tuning datasets and the
  90-configuration grid. (a) Marginal effect of each parameter on mean
  PR-AUC, parameter ranges mapped to a common axis: the anchor temperature
  $\alpha$ dominates, while the kernel parameters $q$ and $\beta$ are flat.
  (b) The $\alpha$ effect per dataset, centred: three datasets prefer
  interior-seeking synthesis ($\alpha>0$) and two prefer the boundary-seeking
  regime ($\alpha<0$).}
  \label{fig:sensitivity}
\end{figure*}

Figure~\ref{fig:sensitivity}(a) shows the marginal effect of each hyperparameter on mean PR-AUC, with each parameter's range mapped to a common axis. The four parameters separate sharply into one influential variable and three inactive ones. The anchor temperature $\alpha$ rises monotonically across its range, moving from $0.5481$ at $\alpha=-1$ to $0.5543$ at $\alpha=+2$, which represent a span of $0.0062$. The neighbourhood size $k$ contributes a smaller span of $0.0025$. In contrast, the kernel parameters $q$ and $\beta$ remain virtually flat, spanning $0.0005$ and $0.0004$ respectively. These kernel parameter variations are roughly
an order of magnitude below the effect of $\alpha$ and fall within expected fold-level noise.

This pattern is informative in both directions. The fact that the anchor $\alpha$ emerges as the active parameter matches the underlying theory directly, as it is the only parameter that provably reorders the synthesis distribution (Lemma~\ref{lem:palpha}). Furthermore, its monotone empirical effect on PR-AUC parallels its monotone effect on certified clearance established in Proposition~\ref{prop:monotone}. The flat response of $q$ and $\beta$ indicates that the placement kernel is highly robust rather than delicately tuned. The method does not depend on locating a narrow operating point for
either parameter, which means a practitioner adopting the locked values incurs no sensitivity risk. While this stability is a desirable practical property, but it also limits the empirical leverage of Phase~III. This observation aligns directly with
the ablation findings presented in Sect.~\ref{subsec:ablation}.

Figure~\ref{fig:sensitivity}(b) resolves the $\alpha$ effect per dataset, revealing that the aggregate monotonicity conceals real underlying heterogeneity. Three datasets favor interior-seeking synthesis, led by haberman ($+0.0374$ from $\alpha=-1$ to $\alpha=+2$) and ecoli3 ($+0.0138$). In contrast, two datasets favor the boundary-seeking regime, abalone19 ($-0.0099$) and
glass0 ($-0.0093$), while yeast3 remains virtually indifferent ($-0.0010$). The signed parameter defined in Eq.~\eqref{eq:anchor_dist} therefore does not merely interpolate between two philosophies in theory. Distinct datasets favor distinct operational regimes in practice, demonstrating that restricting the parameter to $\alpha\ge 0$ would leave two of the five datasets underserved.

We also tested whether the certified fraction identifies which regime a dataset prefers, since it is computable before training and would serve as a convenient selection rule. It does not. Across the five tuning datasets, the Spearman correlation between the certified fraction and the $\alpha$ slope is $\rho = 0.000$ with $p = 1.00$. Selecting $\alpha$ from the certificate is therefore not supported by these data, and we do not recommend it. Consequently, the parameter remains one to tune through standard validation. Given $n=5$, this observation is purely descriptive and is not offered as evidence of absence.

\subsection{Computational cost}\label{subsec:runtime}

Sect.~\ref{subsec:algorithm} argued that CISO's cost is dominated by the one-time neighborhood construction. To evaluate this empirically, we report the measured resampling time per fold, separating the two operational regimes in which the cost
behaves differently. The first regime comprises the small KEEL datasets, where the dense-distance path applies. The second regime comprises the scale suite, where the tree-based path applies.

On the evaluation suite, CISO is the most expensive sampler in the comparison. Its median resampling time is $47$\,ms per fold, compared to $1.2$\,ms for SMOTE, $1.9$\,ms for ADASYN, $16$\,ms for Safe-Level-SMOTE, and $26$\,ms for G-SMOTE. This represents a factor of $39$ over SMOTE at the median, rising to roughly $300\times$ on the largest KEEL datasets, such as $0.40$\,s against $1.4$\,ms on segment0. The computational overhead has a clear source. CISO constructs three distinct neighborhood structures, whereas SMOTE constructs only one. Specifically, while SMOTE only requires a minority-to-minority neighborhood, CISO constructs minority-to-majority and majority-internal neighborhoods in addition to the same minority-to-minority structure. CISO requires all three in order to build the density field defined in Eq.~\eqref{eq:density_field}. The density-aware baselines sit between these two extremes. Methods that compute additional geometry, such as Safe-Level-SMOTE and G-SMOTE, cost an order of magnitude more than SMOTE while remaining an order of magnitude below CISO.

\begin{figure*}[!htbp]
\centering
  \includegraphics[width=\textwidth]{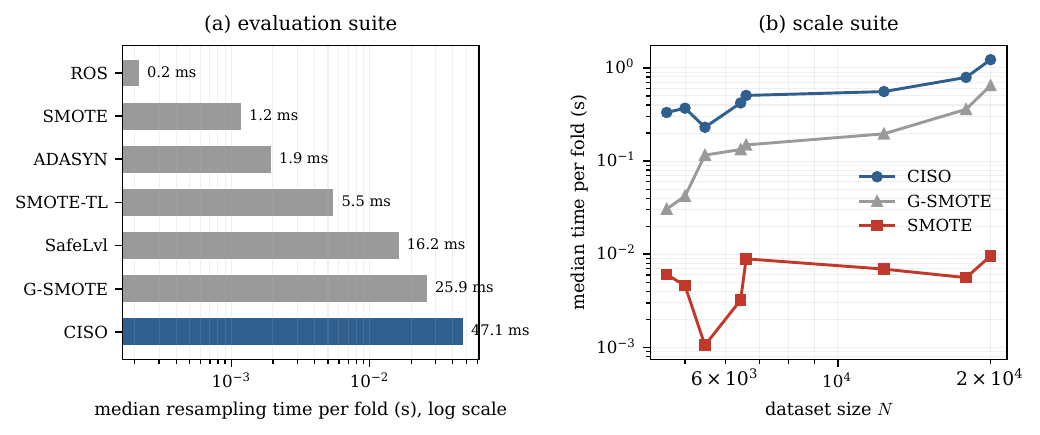}
  \caption{Resampling cost. (a) Median time per fold on the 37 evaluation
  datasets, log scale; CISO is the most expensive sampler, at $39\times$
  SMOTE. (b) Scaling on the scale suite, log--log: with tree-based
  neighbour queries the cost grows near-linearly in $N$, reaching $1.23$\,s
  at $N=20{,}000$. Folds on which a sampler failed are excluded.}
  \label{fig:runtime}
\end{figure*}

In absolute terms these computational costs are small, and the relevant question is how they scale with dataset size. Figure~\ref{fig:runtime}(b) reports the scale suite performance, where CISO employs exact tree-based queries. Median resampling time is $0.23$\,s on page\_blocks ($N=5473$), $0.51$\,s on musk ($N=6598$, $d=166$), $0.79$\,s on HTRU2 ($N=17898$), and $1.23$\,s on letter ($N=20000$). The longest single fold across the scale suite took $1.51$\,s. Between $N\approx5\times10^3$ and $N=2\times10^4$, the measured time grows by roughly a factor of five. This empirical growth aligns with the near-linear $O(Nkd\log N)$ complexity of the tree-based search path, rather than the quadratic cost characteristic of the dense path. Furthermore, the runtime ratio relative to SMOTE narrows significantly in this scale regime, as nearest-neighbour retrieval becomes the dominant
cost for both methods.

We state the computational trade-off plainly. CISO is one to two orders of magnitude slower than SMOTE per invocation, which represents a genuine disadvantage for applications that resample continuously at high frequency. For the ordinary setting, serving as a single preprocessing step prior to classifier training, this cost does not an obstacle. Model fitting itself exceeds one second on all but the smallest datasets, so an absolute cost of $47$\,ms on typical benchmark data and $1.2$\,s at $N=20{,}000$ is well within operational bounds. Table~\ref{tab:rq1} and Fig.~\ref{fig:runtime} should be read together. The additional geometry CISO computes secures the guarantees detailed in Sect.~\ref{subsec:guarantees} alongside the empirical reliability shown in Sect.~\ref{subsec:reliability}, all at a computational cost that remains measurable yet modest relative to the surrounding training pipeline.

\subsection{Reliability}\label{subsec:reliability}

Predictive evaluation presupposes that a method consistently generates output, which is not guaranteed in practice. Samplers that rely on auxiliary geometric structures can encounter input distributions for which those structures cannot be constructed. The experimental protocol explicitly records such failures rather than resolving them through ad hoc fallbacks (Sect.~\ref{subsec:baselines}).

Across the 37 evaluation datasets, nine of the ten evaluable methods, along with all four CISO variants, completed every fold. Two competing baselines experienced execution failures. ADASYN failed on $0.9\%$ of folds across two datasets, occuring when a minority instance had no majority neighbor and its adaptive ratio became undefined. KMeans-SMOTE failed on $50.3\%$ of folds spanning 23 datasets, These errors were caused by a \texttt{RuntimeError} triggered when no cluster contained a sufficient proportion of minority instances. This behavior represents the library's internal safeguard against synthesizing from clusters it judges unreliable. Consequently, this outcome reflects a genuine algorithmic limitation of the method rather than an implementation defect.

These failures follow a systematic pattern rather than a random distribution. The failure rate of KMeans-SMOTE's correlates strongly with the dataset imbalance ratio (IR), as evidenced by a Spearman coefficient of $\rho=+0.61$ ($p=0.0001$). On the 18 datasets with an IR below $10$, the mean failure rate measures $25\%$. This rate rises to $65\%$ across the 14 datasets with an IR between $10$ and $25$. The method reaches complete execution failure across all five datasets with an IR exceeding $25$. Consequently, the method becomes unavailable in direct proportion to the severity of the imbalance it is designed to
address. As the minority class thins, the foundational assumption that it forms identifiable and sufficiently balanced clusters ceases to hold.

CISO and its four variants completed $11{,}100$ fold-level evaluations on the evaluation suite and $360$ on the scale suite without a single failure. This execution stability holds across imbalance ratios from $1.8$ to $129$, training folds containing as few as six minority instances, and feature dimensionality upto $166$. This robustness stems from structural design rather than favorable conditions. Lemma~\ref{lem:phase2}(i) establishes that every Phase-II weight is strictly positive, ensuring that the selection distribution is well-defined for every anchor without degenerate branches. Furthermore, Lemma~\ref{lem:palpha}(iii) guarantees that every anchor retains a positive seeding probability, while Lemma~\ref{lem:placement}(i) ensures that the placement density is well-defined on $[0,1]$ for all admissible parameters. These theoretical guarantees hold irrespective of certification status. For example, on \texttt{abalone19}, where no anchor satisfies $\gamma_i>0$, CISO still generates the full complement of $\Delta N$ instances (Sect.~\ref{subsec:algorithm}).

We do not claim reliability as an exclusive property of CISO. SMOTE, ROS, and six other baseline methods demonstrate equal dependability in these experiments. What distinguishes CISO within this set is that its well-posedness is mathematically established rather than merely empirically observed. Consequently, this operational stability holds by construction for arbitrary inputs configurations beyond the tested benchmark suite.

\section{Discussion}\label{sec:discussion}

\paragraph{What certification provides}\label{subsec:disc-certification}

CISO centers on a foundational certification objective. Specifically, it generates synthetic minority instances that carry a per-instance safety property established strictly by construction rather than heuristic design. Section~\ref{sec:results} evaluated whether this theoretical objective could be achieved without sacrificing predictive quality. The empirical findings confirm that it can. CISO is statistically equivalent to SMOTE, ranks second of eleven under gradient boosting, and significantly outperforms two established baselines. Crucially, every synthetic instance seeded by a certified anchor carries a provable distance guarantee from the majority class that no competing method in the comparison provides. Certification is therefore obtained alongside competitive predictive performance, rather than in place of it.

The primary value of this property lies in what a computed guarantee enables downstream. A sampler that reports per instance distance guarantees supplies the explicit evidence required by auditing and certification regimes in regulated domains~\citep{ci7}. Importantly, it provides this verification before any synthetic instance is generated rather than through post-hoc inspection of a black-box generator \citep{ci4}. Execution reliability follows from this same structural construction. CISO completed all $11{,}460$ fold-level evaluations without a single failure (Sect.~\ref{subsec:reliability}). Because this stability stems directly from the positivity and well-posedness results established in Sect.~\ref{sec:method} rather than from favourable test conditions, it extends reliably to input configurations beyond those evaluated here. Where synthetic data feeds downstream models, human reviewer workflows, or regulatory audit trails, these structural guarantees determine whether the data can be deployed safely. They represent precisely the operational properties that standard predictive metrics fail to measure.

\paragraph{The fidelity-safety frontier}\label{subsec:disc-frontier}

CISO's safety guarantees extend beyond purely formal derivations (Sect.\ref{subsec:fidelity}). The $\alpha$ parameter sweep moves continuously through the fidelity-clearance plane. Clearance rises monotonically in exact accordance with
Proposition~\ref{prop:monotone}, while held-out fidelity reaches its trough at $\alpha=0$ precisely as predicted by the reweighting interpretation of Eq.~\eqref{eq:anchor_dist}. No competing method evaluated in this study offers comparable continuity. Instead, each baseline method occupies a single fixed point within the space, accessible only by switching to an entirely different algorithmic implementation. The practical implication is that $\alpha$ functions as a single control hyperparameter balancing two distinct failure modes previously treated as separate design philosophies in the literature. These failure modes comprise distributional drift toward the majority decision boundary, as seen in ADASYN and Borderline-SMOTE, alongside drift into the minority interior away from the true class-conditional density, as seen in CISO's locked configuration. Neither extreme is without operational cost. Figure~\ref{fig:frontier} renders these trade-offs transparent within a unified framework, eliminating the need for practitioners to compare disparate algorithms to evaluate this balance.

The empirical analysis also delineates the operational boundaries of the observed frontier. Notably, these findings do not assert that CISO Pareto-dominates established baseline methods across the entire objective space. When evaluates at equivalent clearance threshold, both ROS and SMOTE maintain higher fidelity efficiency than any individual point evaluated along the CISO parameter sweep. Consequently, CISO's principal contribution lies in establishing the existence and mathematical continuity of this trade-off curve, rather than claiming spatial dominance over static alternatives.

\paragraph{The Predictive Ceiling of Geometric Refinement}\label{subsec:disc-ceiling}

One might expect that instances provably distant from the majority class would also yield better classification performance. They do not, because the two objectives diverge past a certain threshold. Classification accuracy relies heavily on information regarding the location of the decision boundary, which is concentrated near the boundary itself. An instance placed deep in the minority interior is maximally safe and minimally informative, merely adding density to a region already well-described by the original data. This phenomenon corresponds directly to the mechanism measured in our fidelity analysis (Sect.~\ref{subsec:fidelity}) where reweighting synthesis toward safety shifts the true class-conditional density. Consequently, safety and informativeness represent conflicting objective past a given point, and CISO's fixed configuration intentionally operates at the safe end of this spectrum.

CISO's competitive standing (Sect.~\ref{subsec:rq1}) sits within a field that is itself narrowly dispersed, and that dispersion is the more consequential finding. The fact that CISO does not predictively dominate SMOTE is informative in its own right. CISO is constructed to be a maximally principled instance of geometric refinement, where every design choice derives from a stated criterion (Sect.~\ref{sec:method}) rather than empirical benchmark fitting. Furthermore, it establishes the principle that anchor-level information cancels identically when weights are normalized within a single anchor's neighbourhood (Lemma~\ref{lem:cancellation}), meaning such information must operate during anchor selection. That a model built this deliberately still fails to achieve statistical separation from SMOTE on PR-AUC provides evidence about the underlying paradigm rather than specific method. Indeed, no individual phase is independently detectable on that metric (Sect.~\ref{subsec:ablation}), and no method in the comparison, including ours, consistently outperforms class weighting or random oversampling. The predictive headroom available to geometric refinement of interpolation oversampling appears close to exhausted at the point SMOTE already occupies. This observation is consistent with recent large-scale findings that sophisticated resampling rarely outperforms simple baselines under strong classifiers \citep{ci14,ci15}.

We therefore make no claim that CISO should replace SMOTE on predictive grounds. The main empirical results support such a claim only against the two baselines CISO significantly outperforms, while the runtime analysis weighs directly against it wherever resampling speed is the primary operational constraint. What these results support instead is a precise claim about the underlying design space. When a paradigm's predictive headroom is this narrow, the axis on
which methods can still be meaningfully distinguished is not accuracy but the geometry of what they generate, and whether that geometry can be stated as a guarantee rather than an intention. This is the axis on which CISO is designed to differ, and the one the preceding subsections measure.

\paragraph{When certification helps, and when it does not}\label{subsec:disc-when}

The certificate tracks a specific and verifiable quantity, measuring the exact fraction of the minority class that admits a distance guarantee at a chosen $k$. As demonstrated in Figure~\ref{fig:certificate}, the metric responds directly to minority geometry by separating safe interior clusters from bridging small disjuncts within a single dataset. Interpreted as a structural descriptor, the certified fraction characterize the underlying difficulty of the data prior to classifier training. Consequently, this measure directly determines whether a given synthetic instance maintains a proven distance from the majority class.

Two things it does not do are worth stating precisely, because both were tested emperically. The certified fraction does not predict CISO's advantage over SMOTE, as the correlation failed to reach statistical significance across all evaluated classifier. Furthermore, it does not predict which $\alpha$ regime a dataset prefers. Across the five tuning datasets, that correlation measured exactly zero. We report both null results without qualification because together they delimit the scope of the certificate. Specifically, the certificate is a statement about the worst-case geometric safety of individual synthetic instances, proved unconditionally from $c_i$ and $r_i$ (Lemma~\ref{lem:margin}). It is not a statement about aggregate downstream performance, which depends on the classifier, the decision boundary, and interactions between synthesis and the learning algorithm that the certificate does not model. The certificate governs the safety of what is generated rather than accuracy of what is subsequently learned from it. Interpreting the certified fraction as a predictive performance forecast is therefore unsupported.



\section{Conclusion}\label{sec:conclusion}

This study focuses on a certification objective, generating samples that possess a per-instance safety property established by construction, while remaining competitive on predictive performance. It introduces CISO, which provides a three-phase interpolation framework carrying theoretical guarantees that prior interpolation oversamplers lack. These include a per-instance certified clearance from the majority class for sufficiently safe anchors, a single signed hyperparameter that provably and monotonically governs the balance between boundary-seeking and interior-seeking synthesis, and operational well-posedness that eliminates degenerate algorithmic branches. Its architecture follows a structural principle established in this work. Specifically, safety information that characterizes an anchor rather than its individual neighbors must govern which anchor seeds synthesis. This restriction holds because anchor selection is the only decision stage across which such information varies.

The empirical evidence, gathered under a protocol preregistered prior to execution across 45 datasets, four classifiers, and eleven competing methods, establishes four properties of the framework. First, CISO demonstrates statistical equivalence to SMOTE and to the strongest members of its algorithmic family on the primary threshold-free metric. It ranks second among eleven under gradient boosting and performs significantly ahead of two density-aware baselines, at approximately $39\times$ the resampling cost of SMOTE. Second, ablating each phase measurably alters the geometry of synthesis, indicating that the components govern placement and safety rather than predictive performance. Third, its distributional fidelity to held-out data occupies a mid-field position, trading directly against the certified clearance governed by its control parameter; the temperature parameter traces a continuous, provably monotone path through this fidelity-safety trade-off space, whereas static competing methods occupy only isolated points within it. Fourth, CISO completed all $11{,}460$ fold-level evaluations across both benchmark suites without failure, while a competing baseline failed on half of the evaluation suite with a failure pattern that worsens systematically with increasing class imbalance.

Several directions extend this work. Adaptively setting the temperature parameter based on lightweight data properties would make the fidelity-safety trade-off to specific datasets. This direction is particularly relevant given our finding that the certified fraction alone does not predict the optimal regime. Generalizing the anchor and neighbour distributions to multiclass settings would extend the certification framework beyond binary imbalance. A decision-theoretic framework offers a practical optimization path. If an application can qualify both the cost of an unsafe synthetic instance and the cost of distributionally unfaithfulness, the measured fidelity-safety frontier determines the optimal parameter setting directly. Ultimately, the framework demonstrates that geometric refinement of interpolation oversampling can be rendered provably safe, structurally reliable and precisely characterized for its underlying certification objective, all while maintaining a modest computational cost for standard preprocessing workflows.

\bibliography{sn-bibliography}


\begin{thebibliography}{51}
\ifx \bisbn   \undefined \def \bisbn  #1{ISBN #1}\fi
\ifx \binits  \undefined \def \binits#1{#1}\fi
\ifx \bauthor  \undefined \def \bauthor#1{#1}\fi
\ifx \batitle  \undefined \def \batitle#1{#1}\fi
\ifx \bjtitle  \undefined \def \bjtitle#1{#1}\fi
\ifx \bvolume  \undefined \def \bvolume#1{\textbf{#1}}\fi
\ifx \byear  \undefined \def \byear#1{#1}\fi
\ifx \bissue  \undefined \def \bissue#1{#1}\fi
\ifx \bfpage  \undefined \def \bfpage#1{#1}\fi
\ifx \blpage  \undefined \def \blpage #1{#1}\fi
\ifx \burl  \undefined \def \burl#1{\textsf{#1}}\fi
\ifx \doiurl  \undefined \def \doiurl#1{\url{https://doi.org/#1}}\fi
\ifx \betal  \undefined \def \betal{\textit{et al.}}\fi
\ifx \binstitute  \undefined \def \binstitute#1{#1}\fi
\ifx \binstitutionaled  \undefined \def \binstitutionaled#1{#1}\fi
\ifx \bctitle  \undefined \def \bctitle#1{#1}\fi
\ifx \beditor  \undefined \def \beditor#1{#1}\fi
\ifx \bpublisher  \undefined \def \bpublisher#1{#1}\fi
\ifx \bbtitle  \undefined \def \bbtitle#1{#1}\fi
\ifx \bedition  \undefined \def \bedition#1{#1}\fi
\ifx \bseriesno  \undefined \def \bseriesno#1{#1}\fi
\ifx \blocation  \undefined \def \blocation#1{#1}\fi
\ifx \bsertitle  \undefined \def \bsertitle#1{#1}\fi
\ifx \bsnm \undefined \def \bsnm#1{#1}\fi
\ifx \bsuffix \undefined \def \bsuffix#1{#1}\fi
\ifx \bparticle \undefined \def \bparticle#1{#1}\fi
\ifx \barticle \undefined \def \barticle#1{#1}\fi
\bibcommenthead
\ifx \bconfdate \undefined \def \bconfdate #1{#1}\fi
\ifx \botherref \undefined \def \botherref #1{#1}\fi
\ifx \url \undefined \def \url#1{\textsf{#1}}\fi
\ifx \bchapter \undefined \def \bchapter#1{#1}\fi
\ifx \bbook \undefined \def \bbook#1{#1}\fi
\ifx \bcomment \undefined \def \bcomment#1{#1}\fi
\ifx \oauthor \undefined \def \oauthor#1{#1}\fi
\ifx \citeauthoryear \undefined \def \citeauthoryear#1{#1}\fi
\ifx \endbibitem  \undefined \def \endbibitem {}\fi
\ifx \bconflocation  \undefined \def \bconflocation#1{#1}\fi
\ifx \arxivurl  \undefined \def \arxivurl#1{\textsf{#1}}\fi
\csname PreBibitemsHook\endcsname

\bibitem[\protect\citeauthoryear{Alaa et~al.}{2022}]{ci4}
\begin{bchapter}
\bauthor{\bsnm{Alaa}, \binits{A.}},
\bauthor{\bsnm{Breugel}, \binits{B.V.}},
\bauthor{\bsnm{Saveliev}, \binits{E.S.}},
\bauthor{\bsnm{Schaar}, \binits{M.v.d.}}:
\bctitle{How {Faithful} is your {Synthetic} {Data}? {Sample}-level {Metrics} for {Evaluating} and {Auditing} {Generative} {Models}}.
In: \bbtitle{Proceedings of the 39th {International} {Conference} on {Machine} {Learning}},
vol. \bseriesno{162},
pp. \bfpage{290}--\blpage{306}.
\bpublisher{PMLR},
\blocation{Baltimore}
(\byear{2022}).
\burl{https://proceedings.mlr.press/v162/alaa22a.html}
\end{bchapter}
\endbibitem

\bibitem[\protect\citeauthoryear{Alcala-Fdez et~al.}{}]{EC8}
\begin{botherref}
\oauthor{\bsnm{Alcala-Fdez}, \binits{J.}},
\oauthor{\bsnm{Fernández}, \binits{A.}},
\oauthor{\bsnm{Luengo}, \binits{J.}},
\oauthor{\bsnm{Derrac}, \binits{J.}},
\oauthor{\bsnm{García}, \binits{S.}},
\oauthor{\bsnm{Sánchez}, \binits{L.}},
\oauthor{\bsnm{Herrera}, \binits{F.}}:
Alcalá-{Fdez}, {J}., et al. (2011) {KEEL} {Data}-{Mining} {Software} {Tool} {Data} {Set} {Repository}. {Integration} of {Algorithms} and {Experimental} {Analysis} {Framework}. {Journal} of {Multiple}-{Valued} {Logic} and {Soft} {Computing}, 17, 255-287. - {References} - {Scientific} {Research} {Publishing}.
\url{https://www.scirp.org/reference/referencespapers?referenceid=1222176}
\end{botherref}
\endbibitem

\bibitem[\protect\citeauthoryear{Barua et~al.}{2014}]{ci11}
\begin{barticle}
\bauthor{\bsnm{Barua}, \binits{S.}},
\bauthor{\bsnm{Islam}, \binits{M.M.}},
\bauthor{\bsnm{Yao}, \binits{X.}},
\bauthor{\bsnm{Murase}, \binits{K.}}:
\batitle{{MWMOTE}–{Majority} {Weighted} {Minority} {Oversampling} {Technique} for {Imbalanced} {Data} {Set} {Learning}}.
\bjtitle{IEEE Transactions on Knowledge and Data Engineering}
\bvolume{26}(\bissue{2}),
\bfpage{405}--\blpage{425}
(\byear{2014})
\doiurl{10.1109/TKDE.2012.232}
\end{barticle}
\endbibitem

\bibitem[\protect\citeauthoryear{Batista et~al.}{2004}]{ci28}
\begin{barticle}
\bauthor{\bsnm{Batista}, \binits{G.E.A.P.A.}},
\bauthor{\bsnm{Prati}, \binits{R.C.}},
\bauthor{\bsnm{Monard}, \binits{M.C.}}:
\batitle{A study of the behavior of several methods for balancing machine learning training data}.
\bjtitle{ACM SIGKDD Explorations Newsletter}
\bvolume{6}(\bissue{1}),
\bfpage{20}--\blpage{29}
(\byear{2004})
\doiurl{10.1145/1007730.1007735}
\end{barticle}
\endbibitem

\bibitem[\protect\citeauthoryear{Bunkhumpornpat et~al.}{2009}]{ci10}
\begin{bchapter}
\bauthor{\bsnm{Bunkhumpornpat}, \binits{C.}},
\bauthor{\bsnm{Sinapiromsaran}, \binits{K.}},
\bauthor{\bsnm{Lursinsap}, \binits{C.}}:
\bctitle{Safe-{Level}-{SMOTE}: {Safe}-{Level}-{Synthetic} {Minority} {Over}-{Sampling} {TEchnique} for {Handling} the {Class} {Imbalanced} {Problem}}.
In: \beditor{\bsnm{Theeramunkong}, \binits{T.}},
\beditor{\bsnm{Kijsirikul}, \binits{B.}},
\beditor{\bsnm{Cercone}, \binits{N.}},
\beditor{\bsnm{Ho}, \binits{T.-B.}} (eds.)
\bbtitle{Advances in {Knowledge} {Discovery} and {Data} {Mining}},
pp. \bfpage{475}--\blpage{482}.
\bpublisher{Springer},
\blocation{Berlin, Heidelberg}
(\byear{2009}).
\doiurl{10.1007/978-3-642-01307-2_43}
\end{bchapter}
\endbibitem

\bibitem[\protect\citeauthoryear{Chawla et~al.}{2002}]{ci3}
\begin{barticle}
\bauthor{\bsnm{Chawla}, \binits{N.V.}},
\bauthor{\bsnm{Bowyer}, \binits{K.W.}},
\bauthor{\bsnm{Hall}, \binits{L.O.}},
\bauthor{\bsnm{Kegelmeyer}, \binits{W.P.}}:
\batitle{{SMOTE}: {Synthetic} {Minority} {Over}-sampling {Technique}}.
\bjtitle{Journal of Artificial Intelligence Research}
\bvolume{16},
\bfpage{321}--\blpage{357}
(\byear{2002})
\doiurl{10.1613/jair.953}
\end{barticle}
\endbibitem

\bibitem[\protect\citeauthoryear{Davis}{2006}]{ci32}
\begin{barticle}
\bauthor{\bsnm{Davis}, \binits{H.A.}}:
\batitle{Exploring the {Contexts} of {Relationship} {Quality} between {Middle} {School} {Students} and {Teachers}}.
\bjtitle{The Elementary School Journal}
\bvolume{106}(\bissue{3}),
\bfpage{193}--\blpage{223}
(\byear{2006})
\doiurl{10.1086/501483}
\end{barticle}
\endbibitem

\bibitem[\protect\citeauthoryear{Douzas and Bacao}{2019}]{ci30}
\begin{barticle}
\bauthor{\bsnm{Douzas}, \binits{G.}},
\bauthor{\bsnm{Bacao}, \binits{F.}}:
\batitle{Geometric {SMOTE} a geometrically enhanced drop-in replacement for {SMOTE}}.
\bjtitle{Information Sciences}
\bvolume{501},
\bfpage{118}--\blpage{135}
(\byear{2019})
\doiurl{10.1016/j.ins.2019.06.007}
\end{barticle}
\endbibitem

\bibitem[\protect\citeauthoryear{Douzas et~al.}{2018}]{ci29}
\begin{barticle}
\bauthor{\bsnm{Douzas}, \binits{G.}},
\bauthor{\bsnm{Bacao}, \binits{F.}},
\bauthor{\bsnm{Last}, \binits{F.}}:
\batitle{Improving imbalanced learning through a heuristic oversampling method based on k-means and {SMOTE}}.
\bjtitle{Information Sciences}
\bvolume{465},
\bfpage{1}--\blpage{20}
(\byear{2018})
\doiurl{10.1016/j.ins.2018.06.056}
\end{barticle}
\endbibitem

\bibitem[\protect\citeauthoryear{Demšar}{2006}]{ci33}
\begin{barticle}
\bauthor{\bsnm{Demšar}, \binits{J.}}:
\batitle{Statistical {Comparisons} of {Classifiers} over {Multiple} {Data} {Sets}}.
\bjtitle{Journal of Machine Learning Research}
\bvolume{7}(\bissue{1}),
\bfpage{1}--\blpage{30}
(\byear{2006})
\end{barticle}
\endbibitem

\bibitem[\protect\citeauthoryear{Dua and Graff.}{}]{ci25}
\begin{botherref}
\oauthor{\bsnm{Dua}, \binits{D.}},
\oauthor{\bsnm{Graff.}, \binits{C.}}:
{UCI} {Machine} {Learning} {Repository} {\textbar} {BibSonomy}.
\url{https://www.bibsonomy.org/bibtex/d8e9576e59062411ac69a6a57d8da4fd}
\end{botherref}
\endbibitem

\bibitem[\protect\citeauthoryear{Elreedy and Atiya}{2019}]{ci19}
\begin{barticle}
\bauthor{\bsnm{Elreedy}, \binits{D.}},
\bauthor{\bsnm{Atiya}, \binits{A.F.}}:
\batitle{A {Comprehensive} {Analysis} of {Synthetic} {Minority} {Oversampling} {Technique} ({SMOTE}) for handling class imbalance}.
\bjtitle{Information Sciences}
\bvolume{505},
\bfpage{32}--\blpage{64}
(\byear{2019})
\doiurl{10.1016/j.ins.2019.07.070}
\end{barticle}
\endbibitem

\bibitem[\protect\citeauthoryear{Elor and Averbuch-Elor}{2022}]{ci14}
\begin{botherref}
\oauthor{\bsnm{Elor}, \binits{Y.}},
\oauthor{\bsnm{Averbuch-Elor}, \binits{H.}}:
To {SMOTE}, or not to {SMOTE}?
arXiv.
arXiv:2201.08528 [cs.LG]
(2022).
\doiurl{10.48550/arXiv.2201.08528}
\end{botherref}
\endbibitem

\bibitem[\protect\citeauthoryear{Elreedy et~al.}{2024}]{ci20}
\begin{barticle}
\bauthor{\bsnm{Elreedy}, \binits{D.}},
\bauthor{\bsnm{Atiya}, \binits{A.F.}},
\bauthor{\bsnm{Kamalov}, \binits{F.}}:
\batitle{A theoretical distribution analysis of synthetic minority oversampling technique ({SMOTE}) for imbalanced learning}.
\bjtitle{Machine Learning}
\bvolume{113}(\bissue{7}),
\bfpage{4903}--\blpage{4923}
(\byear{2024})
\doiurl{10.1007/s10994-022-06296-4}
\end{barticle}
\endbibitem

\bibitem[\protect\citeauthoryear{Fontanari et~al.}{2022}]{c34}
\begin{bchapter}
\bauthor{\bsnm{Fontanari}, \binits{T.}},
\bauthor{\bsnm{Fróes}, \binits{T.C.}},
\bauthor{\bsnm{Recamonde-Mendoza}, \binits{M.}}:
\bctitle{Cross-validation {Strategies} for {Balanced} and {Imbalanced} {Datasets}}.
In: \beditor{\bsnm{Xavier-Junior}, \binits{J.C.}},
\beditor{\bsnm{Rios}, \binits{R.A.}} (eds.)
\bbtitle{Intelligent {Systems}},
pp. \bfpage{626}--\blpage{640}.
\bpublisher{Springer},
\blocation{Cham}
(\byear{2022}).
\doiurl{10.1007/978-3-031-21686-2_43}
\end{bchapter}
\endbibitem

\bibitem[\protect\citeauthoryear{Fernandez et~al.}{2018}]{ci1}
\begin{barticle}
\bauthor{\bsnm{Fernandez}, \binits{A.}},
\bauthor{\bsnm{Garcia}, \binits{S.}},
\bauthor{\bsnm{Herrera}, \binits{F.}},
\bauthor{\bsnm{Chawla}, \binits{N.V.}}:
\batitle{{SMOTE} for {Learning} from {Imbalanced} {Data}: {Progress} and {Challenges}, {Marking} the 15-year {Anniversary}}.
\bjtitle{Journal of Artificial Intelligence Research}
\bvolume{61},
\bfpage{863}--\blpage{905}
(\byear{2018})
\doiurl{10.1613/jair.1.11192}
\end{barticle}
\endbibitem

\bibitem[\protect\citeauthoryear{Gao et~al.}{2014}]{ci21}
\begin{barticle}
\bauthor{\bsnm{Gao}, \binits{M.}},
\bauthor{\bsnm{Hong}, \binits{X.}},
\bauthor{\bsnm{Chen}, \binits{S.}},
\bauthor{\bsnm{Harris}, \binits{C.J.}},
\bauthor{\bsnm{Khalaf}, \binits{E.}}:
\batitle{{PDFOS}: {PDF} estimation based over-sampling for imbalanced two-class problems}.
\bjtitle{Neurocomputing}
\bvolume{138},
\bfpage{248}--\blpage{259}
(\byear{2014})
\doiurl{10.1016/j.neucom.2014.02.006}
\end{barticle}
\endbibitem

\bibitem[\protect\citeauthoryear{Gosain and Sardana}{2017}]{q11}
\begin{bchapter}
\bauthor{\bsnm{Gosain}, \binits{A.}},
\bauthor{\bsnm{Sardana}, \binits{S.}}:
\bctitle{Handling class imbalance problem using oversampling techniques: {A} review}.
In: \bbtitle{2017 {International} {Conference} on {Advances} in {Computing}, {Communications} and {Informatics} ({ICACCI})},
pp. \bfpage{79}--\blpage{85}
(\byear{2017}).
\doiurl{10.1109/ICACCI.2017.8125820}
\end{bchapter}
\endbibitem

\bibitem[\protect\citeauthoryear{He et~al.}{2008}]{ci8}
\begin{bchapter}
\bauthor{\bsnm{He}, \binits{H.}},
\bauthor{\bsnm{Bai}, \binits{Y.}},
\bauthor{\bsnm{Garcia}, \binits{E.A.}},
\bauthor{\bsnm{Li}, \binits{S.}}:
\bctitle{{ADASYN}: {Adaptive} synthetic sampling approach for imbalanced learning}.
In: \bbtitle{2008 {IEEE} {International} {Joint} {Conference} on {Neural} {Networks} ({IEEE} {World} {Congress} on {Computational} {Intelligence})},
pp. \bfpage{1322}--\blpage{1328}
(\byear{2008}).
\doiurl{10.1109/IJCNN.2008.4633969} .
\bcomment{ISSN: 2161-4407}
\end{bchapter}
\endbibitem

\bibitem[\protect\citeauthoryear{Han et~al.}{2005}]{ci9}
\begin{bchapter}
\bauthor{\bsnm{Han}, \binits{H.}},
\bauthor{\bsnm{Wang}, \binits{W.-Y.}},
\bauthor{\bsnm{Mao}, \binits{B.-H.}}:
\bctitle{Borderline-{SMOTE}: {A} {New} {Over}-{Sampling} {Method} in {Imbalanced} {Data} {Sets} {Learning}}.
In: \beditor{\bsnm{Huang}, \binits{D.-S.}},
\beditor{\bsnm{Zhang}, \binits{X.-P.}},
\beditor{\bsnm{Huang}, \binits{G.-B.}} (eds.)
\bbtitle{Advances in {Intelligent} {Computing}},
pp. \bfpage{878}--\blpage{887}.
\bpublisher{Springer},
\blocation{Berlin, Heidelberg}
(\byear{2005}).
\doiurl{10.1007/11538059_91}
\end{bchapter}
\endbibitem

\bibitem[\protect\citeauthoryear{Haixiang et~al.}{2017}]{ci2}
\begin{barticle}
\bauthor{\bsnm{Haixiang}, \binits{G.}},
\bauthor{\bsnm{Yijing}, \binits{L.}},
\bauthor{\bsnm{Shang}, \binits{J.}},
\bauthor{\bsnm{Mingyun}, \binits{G.}},
\bauthor{\bsnm{Yuanyue}, \binits{H.}},
\bauthor{\bsnm{Bing}, \binits{G.}}:
\batitle{Learning from class-imbalanced data: {Review} of methods and applications}.
\bjtitle{Expert Systems with Applications}
\bvolume{73},
\bfpage{220}--\blpage{239}
(\byear{2017})
\doiurl{10.1016/j.eswa.2016.12.035}
\end{barticle}
\endbibitem

\bibitem[\protect\citeauthoryear{Kulkarni et~al.}{2020}]{q6}
\begin{bchapter}
\bauthor{\bsnm{Kulkarni}, \binits{A.}},
\bauthor{\bsnm{Chong}, \binits{D.}},
\bauthor{\bsnm{Batarseh}, \binits{F.A.}}:
\bctitle{5 - {Foundations} of data imbalance and solutions for a data democracy}.
In: \beditor{\bsnm{Batarseh}, \binits{F.A.}},
\beditor{\bsnm{Yang}, \binits{R.}} (eds.)
\bbtitle{Data {Democracy}},
pp. \bfpage{83}--\blpage{106}.
\bpublisher{Academic Press}, \blocation{???}
(\byear{2020}).
\burl{https://doi.org/10.1016/B978-0-12-818366-3.00005-8}
\end{bchapter}
\endbibitem

\bibitem[\protect\citeauthoryear{Koziarski}{2021}]{q23}
\begin{bchapter}
\bauthor{\bsnm{Koziarski}, \binits{M.}}:
\bctitle{{CSMOUTE}: {Combined} {Synthetic} {Oversampling} and {Undersampling} {Technique} for {Imbalanced} {Data} {Classification}}.
In: \bbtitle{2021 {International} {Joint} {Conference} on {Neural} {Networks} ({IJCNN})},
pp. \bfpage{1}--\blpage{8}
(\byear{2021}).
\doiurl{10.1109/IJCNN52387.2021.9533415}
\end{bchapter}
\endbibitem

\bibitem[\protect\citeauthoryear{Krawczyk}{2016}]{q5}
\begin{barticle}
\bauthor{\bsnm{Krawczyk}, \binits{B.}}:
\batitle{Learning from imbalanced data: open challenges and future directions}.
\bjtitle{Progress in Artificial Intelligence}
\bvolume{5}(\bissue{4}),
\bfpage{221}--\blpage{232}
(\byear{2016})
\doiurl{10.1007/s13748-016-0094-0}
\end{barticle}
\endbibitem

\bibitem[\protect\citeauthoryear{Leevy et~al.}{2018}]{q17}
\begin{barticle}
\bauthor{\bsnm{Leevy}, \binits{J.L.}},
\bauthor{\bsnm{Khoshgoftaar}, \binits{T.M.}},
\bauthor{\bsnm{Bauder}, \binits{R.A.}},
\bauthor{\bsnm{Seliya}, \binits{N.}}:
\batitle{A survey on addressing high-class imbalance in big data}.
\bjtitle{Journal of Big Data}
\bvolume{5}(\bissue{1}),
\bfpage{42}
(\byear{2018})
\doiurl{10.1186/s40537-018-0151-6}
\end{barticle}
\endbibitem

\bibitem[\protect\citeauthoryear{Lin and Liang}{2025}]{ci18}
\begin{barticle}
\bauthor{\bsnm{Lin}, \binits{J.}},
\bauthor{\bsnm{Liang}, \binits{L.}}:
\batitle{A non-parameter oversampling approach for imbalanced data classification based on hybrid natural neighbors}.
\bjtitle{Applied Intelligence}
\bvolume{55}(\bissue{5}),
\bfpage{362}
(\byear{2025})
\doiurl{10.1007/s10489-025-06236-4}
\end{barticle}
\endbibitem

\bibitem[\protect\citeauthoryear{Lemaître et~al.}{2017}]{ci31}
\begin{barticle}
\bauthor{\bsnm{Lemaître}, \binits{G.}},
\bauthor{\bsnm{Nogueira}, \binits{F.}},
\bauthor{\bsnm{Aridas}, \binits{C.K.}}:
\batitle{Imbalanced-learn: {A} {Python} {Toolbox} to {Tackle} the {Curse} of {Imbalanced} {Datasets} in {Machine} {Learning}}.
\bjtitle{Journal of Machine Learning Research}
\bvolume{18}(\bissue{17}),
\bfpage{1}--\blpage{5}
(\byear{2017})
\end{barticle}
\endbibitem

\bibitem[\protect\citeauthoryear{Loftsgaarden and Quesenberry}{1965}]{ci24}
\begin{barticle}
\bauthor{\bsnm{Loftsgaarden}, \binits{D.O.}},
\bauthor{\bsnm{Quesenberry}, \binits{C.P.}}:
\batitle{A {Nonparametric} {Estimate} of a {Multivariate} {Density} {Function}}.
\bjtitle{The Annals of Mathematical Statistics}
\bvolume{36}(\bissue{3}),
\bfpage{1049}--\blpage{1051}
(\byear{1965})
\doiurl{10.1214/aoms/1177700079}
\end{barticle}
\endbibitem

\bibitem[\protect\citeauthoryear{Li et~al.}{2021}]{ci17}
\begin{barticle}
\bauthor{\bsnm{Li}, \binits{J.}},
\bauthor{\bsnm{Zhu}, \binits{Q.}},
\bauthor{\bsnm{Wu}, \binits{Q.}},
\bauthor{\bsnm{Fan}, \binits{Z.}}:
\batitle{A novel oversampling technique for class-imbalanced learning based on {SMOTE} and natural neighbors}.
\bjtitle{Information Sciences}
\bvolume{565},
\bfpage{438}--\blpage{455}
(\byear{2021})
\doiurl{10.1016/j.ins.2021.03.041}
\end{barticle}
\endbibitem

\bibitem[\protect\citeauthoryear{Matharaarachchi et~al.}{2024}]{ci27}
\begin{barticle}
\bauthor{\bsnm{Matharaarachchi}, \binits{S.}},
\bauthor{\bsnm{Domaratzki}, \binits{M.}},
\bauthor{\bsnm{Muthukumarana}, \binits{S.}}:
\batitle{Enhancing {SMOTE} for imbalanced data with abnormal minority instances}.
\bjtitle{Machine Learning with Applications}
\bvolume{18},
\bfpage{100597}
(\byear{2024})
\doiurl{10.1016/j.mlwa.2024.100597}
\end{barticle}
\endbibitem

\bibitem[\protect\citeauthoryear{Majzoub and Elgedawy}{2020}]{q13}
\begin{barticle}
\bauthor{\bsnm{Majzoub}, \binits{H.}},
\bauthor{\bsnm{Elgedawy}, \binits{I.}}:
\batitle{{AB}-{SMOTE}: {An} {Affinitive} {Borderline} {SMOTE} {Approach} for {Imbalanced} {Data} {Binary} {Classification}}.
\bjtitle{International Journal of Machine Learning and Computing}
\bvolume{10},
\bfpage{31}--\blpage{37}
(\byear{2020})
\doiurl{10.18178/ijmlc.2020.10.1.894}
\end{barticle}
\endbibitem

\bibitem[\protect\citeauthoryear{Moniz and Monteiro}{2021}]{ci15}
\begin{barticle}
\bauthor{\bsnm{Moniz}, \binits{N.}},
\bauthor{\bsnm{Monteiro}, \binits{H.}}:
\batitle{No {Free} {Lunch} in imbalanced learning}.
\bjtitle{Knowledge-Based Systems}
\bvolume{227},
\bfpage{107222}
(\byear{2021})
\doiurl{10.1016/j.knosys.2021.107222}
\end{barticle}
\endbibitem

\bibitem[\protect\citeauthoryear{Menon et~al.}{2013}]{EC2C}
\begin{bchapter}
\bauthor{\bsnm{Menon}, \binits{A.}},
\bauthor{\bsnm{Narasimhan}, \binits{H.}},
\bauthor{\bsnm{Agarwal}, \binits{S.}},
\bauthor{\bsnm{Chawla}, \binits{S.}}:
\bctitle{On the {Statistical} {Consistency} of {Algorithms} for {Binary} {Classification} under {Class} {Imbalance}}.
In: \bbtitle{Proceedings of the 30th {International} {Conference} on {Machine} {Learning}},
pp. \bfpage{603}--\blpage{611}.
\bpublisher{PMLR},
\blocation{Atlanta}
(\byear{2013}).
\burl{https://proceedings.mlr.press/v28/menon13a.html}
\end{bchapter}
\endbibitem

\bibitem[\protect\citeauthoryear{Mayabadi and Saadatfar}{2022}]{ci13}
\begin{barticle}
\bauthor{\bsnm{Mayabadi}, \binits{S.}},
\bauthor{\bsnm{Saadatfar}, \binits{H.}}:
\batitle{Two density-based sampling approaches for imbalanced and overlapping data}.
\bjtitle{Knowledge-Based Systems}
\bvolume{241},
\bfpage{108217}
(\byear{2022})
\doiurl{10.1016/j.knosys.2022.108217}
\end{barticle}
\endbibitem

\bibitem[\protect\citeauthoryear{Nisevic et~al.}{2025}]{ci7}
\begin{barticle}
\bauthor{\bsnm{Nisevic}, \binits{M.}},
\bauthor{\bsnm{Milojevic}, \binits{D.}},
\bauthor{\bsnm{Spajic}, \binits{D.}}:
\batitle{Synthetic data in medicine: {Legal} and ethical considerations for patient profiling}.
\bjtitle{Computational and Structural Biotechnology Journal}
\bvolume{28},
\bfpage{190}--\blpage{198}
(\byear{2025})
\doiurl{10.1016/j.csbj.2025.05.026}
\end{barticle}
\endbibitem

\bibitem[\protect\citeauthoryear{Napierala and Stefanowski}{2016}]{q15}
\begin{barticle}
\bauthor{\bsnm{Napierala}, \binits{K.}},
\bauthor{\bsnm{Stefanowski}, \binits{J.}}:
\batitle{Types of minority class examples and their influence on learning classifiers from imbalanced data}.
\bjtitle{Journal of Intelligent Information Systems}
\bvolume{46}(\bissue{3}),
\bfpage{563}--\blpage{597}
(\byear{2016})
\doiurl{10.1007/s10844-015-0368-1}
\end{barticle}
\endbibitem

\bibitem[\protect\citeauthoryear{Rao et~al.}{2024}]{q20}
\begin{barticle}
\bauthor{\bsnm{Rao}, \binits{C.}},
\bauthor{\bsnm{Wei}, \binits{X.}},
\bauthor{\bsnm{Xiao}, \binits{X.}},
\bauthor{\bsnm{Shi}, \binits{Y.}},
\bauthor{\bsnm{Goh}, \binits{M.}}:
\batitle{Oversampling method via adaptive double weights and {Gaussian} kernel function for the transformation of unbalanced data in risk assessment of cardiovascular disease}.
\bjtitle{Information Sciences}
\bvolume{665},
\bfpage{120410}
(\byear{2024})
\doiurl{10.1016/j.ins.2024.120410}
\end{barticle}
\endbibitem

\bibitem[\protect\citeauthoryear{Sanguanmak and Hanskunatai}{2016}]{q8}
\begin{bchapter}
\bauthor{\bsnm{Sanguanmak}, \binits{Y.}},
\bauthor{\bsnm{Hanskunatai}, \binits{A.}}:
\bctitle{{DBSM}: {The} combination of {DBSCAN} and {SMOTE} for imbalanced data classification}.
In: \bbtitle{2016 13th {International} {Joint} {Conference} on {Computer} {Science} and {Software} {Engineering} ({JCSSE})},
pp. \bfpage{1}--\blpage{5}
(\byear{2016}).
\doiurl{10.1109/JCSSE.2016.7748928}
\end{bchapter}
\endbibitem

\bibitem[\protect\citeauthoryear{Steinbach and Tan}{2009}]{q36}
\begin{bchapter}
\bauthor{\bsnm{Steinbach}, \binits{M.}},
\bauthor{\bsnm{Tan}, \binits{P.-N.}}:
\bctitle{{kNN}: k-{Nearest} {Neighbors}}.
In: \bbtitle{The {Top} {Ten} {Algorithms} in {Data} {Mining}}.
\bpublisher{Chapman and Hall/CRC},
\blocation{Boca Raton}
(\byear{2009}).
\doiurl{10.1201/9781420089653-15} .
\bcomment{Num Pages: 12}
\end{bchapter}
\endbibitem

\bibitem[\protect\citeauthoryear{Sihag et~al.}{2022}]{kan42}
\begin{bchapter}
\bauthor{\bsnm{Sihag}, \binits{G.}},
\bauthor{\bsnm{Yadav}, \binits{P.}},
\bauthor{\bsnm{Delcroix}, \binits{V.}},
\bauthor{\bsnm{Vijay}, \binits{V.}},
\bauthor{\bsnm{Siebert}, \binits{X.}},
\bauthor{\bsnm{Yadav}, \binits{S.}},
\bauthor{\bsnm{Puisieux}, \binits{F.}}:
\bctitle{Evaluation of {Risk} {Factors} for {Fall} in {Elderly} {People} from {Imbalanced} {Data} using the {Oversampling} {Technique} {SMOTE}:}.
In: \bbtitle{8th {ICT4AWE}},
pp. \bfpage{50}--\blpage{58}.
\bpublisher{SCITEPRESS},
\blocation{Setúbal}
(\byear{2022}).
\doiurl{10.5220/0011041200003188}
\end{bchapter}
\endbibitem

\bibitem[\protect\citeauthoryear{Tanha et~al.}{2020}]{q16}
\begin{barticle}
\bauthor{\bsnm{Tanha}, \binits{J.}},
\bauthor{\bsnm{Abdi}, \binits{Y.}},
\bauthor{\bsnm{Samadi}, \binits{N.}},
\bauthor{\bsnm{Razzaghi}, \binits{N.}},
\bauthor{\bsnm{Asadpour}, \binits{M.}}:
\batitle{Boosting methods for multi-class imbalanced data classification: an experimental review}.
\bjtitle{Journal of Big Data}
\bvolume{7}(\bissue{1}),
\bfpage{70}
(\byear{2020})
\doiurl{10.1186/s40537-020-00349-y}
\end{barticle}
\endbibitem

\bibitem[\protect\citeauthoryear{Tang and He}{2015}]{EC5}
\begin{bchapter}
\bauthor{\bsnm{Tang}, \binits{B.}},
\bauthor{\bsnm{He}, \binits{H.}}:
\bctitle{{KernelADASYN}: {Kernel} {Based} {Adaptive} {Synthetic} {Data} {Generation} for {Imbalanced} {Learning}}.
(\byear{2015}).
\doiurl{https://doi.org/110.1109/CEC.2015.7256954}
\end{bchapter}
\endbibitem

\bibitem[\protect\citeauthoryear{Thabtah et~al.}{2020}]{q4}
\begin{barticle}
\bauthor{\bsnm{Thabtah}, \binits{F.}},
\bauthor{\bsnm{Hammoud}, \binits{S.}},
\bauthor{\bsnm{Kamalov}, \binits{F.}},
\bauthor{\bsnm{Gonsalves}, \binits{A.}}:
\batitle{Data imbalance in classification: {Experimental} evaluation}.
\bjtitle{Information Sciences}
\bvolume{513},
\bfpage{429}--\blpage{441}
(\byear{2020})
\doiurl{10.1016/j.ins.2019.11.004}
\end{barticle}
\endbibitem

\bibitem[\protect\citeauthoryear{Tsallis}{1988}]{ci26}
\begin{barticle}
\bauthor{\bsnm{Tsallis}, \binits{C.}}:
\batitle{Possible generalization of {Boltzmann}-{Gibbs} statistics}.
\bjtitle{Journal of Statistical Physics}
\bvolume{52}(\bissue{1}),
\bfpage{479}--\blpage{487}
(\byear{1988})
\doiurl{10.1007/BF01016429}
\end{barticle}
\endbibitem

\bibitem[\protect\citeauthoryear{Varotto et~al.}{2021}]{ci5}
\begin{botherref}
\oauthor{\bsnm{Varotto}, \binits{G.}},
\oauthor{\bsnm{Susi}, \binits{G.}},
\oauthor{\bsnm{Tassi}, \binits{L.}},
\oauthor{\bsnm{Gozzo}, \binits{F.}},
\oauthor{\bsnm{Franceschetti}, \binits{S.}},
\oauthor{\bsnm{Panzica}, \binits{F.}}:
Comparison of {Resampling} {Techniques} for {Imbalanced} {Datasets} in {Machine} {Learning}: {Application} to {Epileptogenic} {Zone} {Localization} {From} {Interictal} {Intracranial} {EEG} {Recordings} in {Patients} {With} {Focal} {Epilepsy}.
Frontiers in Neuroinformatics
\textbf{15}
(2021)
\doiurl{10.3389/fninf.2021.715421}
\end{botherref}
\endbibitem

\bibitem[\protect\citeauthoryear{Wan et~al.}{2017}]{ci23}
\begin{bchapter}
\bauthor{\bsnm{Wan}, \binits{Z.}},
\bauthor{\bsnm{Zhang}, \binits{Y.}},
\bauthor{\bsnm{He}, \binits{H.}}:
\bctitle{Variational autoencoder based synthetic data generation for imbalanced learning}.
In: \bbtitle{2017 {IEEE} {Symposium} {Series} on {Computational} {Intelligence} ({SSCI})},
pp. \bfpage{1}--\blpage{7}
(\byear{2017}).
\doiurl{10.1109/SSCI.2017.8285168}
\end{bchapter}
\endbibitem

\bibitem[\protect\citeauthoryear{Wang et~al.}{2025}]{q21A}
\begin{barticle}
\bauthor{\bsnm{Wang}, \binits{F.}},
\bauthor{\bsnm{Zheng}, \binits{M.}},
\bauthor{\bsnm{Ma}, \binits{K.}},
\bauthor{\bsnm{Hu}, \binits{X.}}:
\batitle{Resampling approach for imbalanced data classification based on class instance density per feature value intervals}.
\bjtitle{Information Sciences}
\bvolume{692},
\bfpage{121570}
(\byear{2025})
\doiurl{10.1016/j.ins.2024.121570}
\end{barticle}
\endbibitem

\bibitem[\protect\citeauthoryear{Xie et~al.}{2022}]{ci16}
\begin{barticle}
\bauthor{\bsnm{Xie}, \binits{Y.}},
\bauthor{\bsnm{Qiu}, \binits{M.}},
\bauthor{\bsnm{Zhang}, \binits{H.}},
\bauthor{\bsnm{Peng}, \binits{L.}},
\bauthor{\bsnm{Chen}, \binits{Z.}}:
\batitle{Gaussian {Distribution} {Based} {Oversampling} for {Imbalanced} {Data} {Classification}}.
\bjtitle{IEEE Transactions on Knowledge and Data Engineering}
\bvolume{34}(\bissue{2}),
\bfpage{667}--\blpage{679}
(\byear{2022})
\doiurl{10.1109/TKDE.2020.2985965}
\end{barticle}
\endbibitem

\bibitem[\protect\citeauthoryear{Yan et~al.}{2022}]{ci12}
\begin{barticle}
\bauthor{\bsnm{Yan}, \binits{Y.}},
\bauthor{\bsnm{Jiang}, \binits{Y.}},
\bauthor{\bsnm{Zheng}, \binits{Z.}},
\bauthor{\bsnm{Yu}, \binits{C.}},
\bauthor{\bsnm{Zhang}, \binits{Y.}},
\bauthor{\bsnm{Zhang}, \binits{Y.}}:
\batitle{{LDAS}: {Local} density-based adaptive sampling for imbalanced data classification}.
\bjtitle{Expert Systems with Applications}
\bvolume{191},
\bfpage{116213}
(\byear{2022})
\doiurl{10.1016/j.eswa.2021.116213}
\end{barticle}
\endbibitem

\bibitem[\protect\citeauthoryear{Zhang and Li}{2014}]{ci22}
\begin{barticle}
\bauthor{\bsnm{Zhang}, \binits{H.}},
\bauthor{\bsnm{Li}, \binits{M.}}:
\batitle{{RWO}-{Sampling}: {A} random walk over-sampling approach to imbalanced data classification}.
\bjtitle{Information Fusion}
\bvolume{20},
\bfpage{99}--\blpage{116}
(\byear{2014})
\doiurl{10.1016/j.inffus.2013.12.003}
\end{barticle}
\endbibitem

\bibitem[\protect\citeauthoryear{Zafar and Wu}{2026}]{ci6}
\begin{barticle}
\bauthor{\bsnm{Zafar}, \binits{U.}},
\bauthor{\bsnm{Wu}, \binits{F.}}:
\batitle{Methodological challenges in explainable {AI} for fraud detection: a systematic literature review}.
\bjtitle{Artificial Intelligence Review}
\bvolume{59}(\bissue{4}),
\bfpage{115}
(\byear{2026})
\doiurl{10.1007/s10462-026-11516-7}
\end{barticle}
\endbibitem

\end{thebibliography}

\end{document}